%% file: main.tex
\documentclass[acmsmall]{acmart}
  \input{macro}

\AtBeginDocument{%
  \providecommand\BibTeX{{%
    \normalfont B\kern-0.5em{\scshape i\kern-0.25em b}\kern-0.8em\TeX}}}

\setcopyright{acmcopyright}
\copyrightyear{2019}
\acmYear{2019}
\acmDOI{10.1145/1122445.1122456}

\begin{document}


\title{Taking Care of The Discretization Problem}
\subtitle{A Comprehensive Study of the Discretization Problem and A Black-Box Adversarial Attack in Discrete Integer Domain}
\author{Lei Bu}
\email{bulei@nju.edu.cn}
\affiliation{%
  \institution{Nanjing University}
%  \streetaddress{P.O. Box 1212}
  \city{Nanjing}
  \country{China}
 % \postcode{201210}
}
\author{Yuchao Duan}
\email{15050550866@163.com}
\affiliation{%
  \institution{Nanjing University}
%  \streetaddress{P.O. Box 1212}
  \city{Nanjing}
  \country{China}
 % \postcode{201210}
}

\author{Fu Song}
\email{songfu@shanghaitech.edu.cn}
\affiliation{%
  \institution{ShanghaiTech University}
%  \streetaddress{P.O. Box 1212}
  \city{Shanghai}
  \country{China}
 % \postcode{201210}
}

\author{Zhe Zhao}
\email{zhaozhe1@shanghaitech.edu.cn}
\affiliation{%
  \institution{ShanghaiTech University}
 % \streetaddress{1 Th{\o}rv{\"a}ld Circle}
 % \city{Hekla}
   \city{Shanghai}
  \country{China}}
 % \country{Singapore}}

\renewcommand{\shortauthors}{Duan and Zhao, et al.}


\begin{abstract}
Numerous adversarial attacks on neural network based classifiers have been proposed recently with high success rate. Neural network based image classifiers usually normalize valid images into some real continuous domain and make classification decisions on the normalized images using the neural networks. However, existing attacks often craft adversarial examples in such domain,
which may become benign once denormalized back into
the discrete integer domain, known as the discretization problem.
This problem has been mentioned in some work,
but has received relatively little attention.

To understand the impacts of the discretization problem,
in this work, we report the first comprehensive study of existing works on adversarial attacks against neural network based image classification systems.
We theoretically analyze 35 representative methods and empirically study 20 representative open source tools for crafting adversarial images.
%Our study reveals that the discretization problem
%is far more serious than originally thought.
We found 29/35 (theoretically) and 14/20 (empirically), are affected, e.g., the success rate could dramatically drop from 100\% to 10\%.
%For instance, %some black-box methods downgrade to gray-box ones,
%methods having higher success rates drop down to lower success rates using their default parameters, e.g., from $100\%$ to $1\%$,
%and it is difficult to alleviate the discretization problem for many methods and tools.
This reveals that the discretization problem
is far more serious than originally thought and suggests that it should be taken into account seriously
when crafting adversarial examples and measuring attack success rate.

As a first step towards addressing this problem in black-box  scenario, we propose a novel
method which directly crafts adversarial examples in discrete integer domains. Our method reduces adversarial attack problem to a derivative-free optimization (DFO) problem for which we propose a classification model-based DFO algorithm.
%Our method is able to craft adversarial images by derivative-free search in the discrete integer domain.
Experimental results show that our method achieves close to 100\% attack success rates for both targeted
and untargeted attacks, comparable to the most popular white-box methods (FGSM, BIM and C\&W),
and significantly outperforms representative black-box methods (ZOO, AutoZOOM, NES-PGD, Bandits, FD, FD-PSO and GenAttack). % in terms of adversarial examples in the discrete integer domain.
% than representative black-box methods (e.g., ZOO, AutoZOOM, NES-PGD, Bandits and GenAttack). %, no matter white-box or black-box.
Moreover, our method successfully breaks the winner of NIPS 2017 competition on defense with 100\% success rate.
Our results suggest that  discrete optimization algorithms open up a promising area of
research into effective black-box attacks.
\end{abstract}



\maketitle

\input{introduction}

\input{related_work}

%==============================================================================================
\input{background}
\input{discretization}

\input{methodology}

\input{experiments}

\section{Conclusion and Future Work}
\label{sec:concl}
We conducted the first comprehensive study of 35 methods and 20 open source tools
for crafting adversarial examples,  in an attempt to understand the impacts of the discretization problem.
Our study revealed that most of these methods and tools are affected by this problem
and researchers should pay more attention when designing
adversarial example attacks and measuring attack success rate.
We also proposed  strategies to avoid or alleviate
the discretization problem, which can improve TSR of some tools,  at the cost of attack efficiency or imperceptibility of adversarial examples.

We proposed a black-box method by designing a classification model-based derivative-free optimization method.
Our method directly crafts adversarial examples in discrete integer domains, hence does not have the discretization problem and is able to attack a wide range of classifiers including non-differentiable ones.
Our attack method requires access to the probability distribution of classes for each test input and does not rely on the gradient of the objective function, but instead,
learns from samples of the search space. We implemented our method into tool DFA, and conducted an intensive set of experiments on MNIST and ImageNet in both untargeted and targeted scenarios.
The experimental results show that our method achieved close to 100\% attack success rate,
comparable to the white-box methods (FGSM, BIM and C\&W)
and  outperformed the state-of-the-art black-box methods.
Moreover, our method achieved $100\%$ success rate on the winner of NIPS 2017 competition on defense, % success rate
and achieved the same result as the best white-box attack in MNIST Challenge.
Our results suggest that classification model-based derivative-free
discrete optimization opens up a promising research direction into effective
black-box attacks.
Our method could serve as a
test for designing robust networks.

In future, we plan to lift our generic method to other neural network based systems such as face recognition
systems~\cite{SBBR16} and speech recognition~\cite{YuanCZLL0ZH0G18,CarliniMVZSSWZ16}.
It is also worth investigating how to intergrade gradient estimation techniques into our sampling.
This may improves query efficiency of our method.


\begin{acks}
This work is supported by the National Natural Science Foundation of China (NSFC) Grants (Nos. 61532019,
61761136011 and  61572249),
\end{acks}


%\appendix
%\input{appendix}
\appendix
\input{discussNormalization}

\input{discussParameter}
\end{document}

%% file: macro.tex
\usepackage{stmaryrd,dsfont,amsmath,amsfonts}
\usepackage{semantic,multirow,multicol,graphicx}
\usepackage[algoruled,linesnumbered,noend]{algorithm2e}
\usepackage[noend]{algorithmic}
\usepackage{tikz}
\usepackage{pifont}
\usepackage{threeparttable}
\usepackage{colortbl}
\usepackage{lscape}
\usepackage{balance}
\usepackage{tcolorbox}
\newcommand{\tabincell}[2]{\begin{tabular}{@{}#1@{}}#2\end{tabular}}

\definecolor{mygray}{gray}{.9}

\usetikzlibrary{automata,arrows,positioning}

\newcommand{\cmark}{\ding{51}}%
\newcommand{\xmark}{\ding{55}}%
\newcommand{\tool}{{\sc DFA}\xspace}

\newcommand{\Ttop}{{\tt Top}}
\newcommand{\Prr}{{\mathcal{P}}}

\newcommand{\adv}{{\tt adv}}

\usetikzlibrary{shapes.geometric}

\usepackage{hyperref}

%% file: introduction.tex
 \section{Introduction}
\label{sec:intr}
In the past 10 years, machine learning algorithms, fueled by massive amounts of data,
achieve human-level performance or better on a number of tasks.
Models produced by machine learning algorithms, especially deep neural networks,
are increasingly being deployed in a variety of applications %including safety-critical applications,
such as autonomous driving~\cite{Holley18,Apollo,Waymo}, medical diagnostics~\cite{CGGS12,SWS17,PCG15},
speech processing~\cite{HDYDMJSVNSK12},  computer vision~\cite{KTSLSF14,KSH17},
robotics~\cite{ZLMUC15,LPKIQ18},
natural language processing~\cite{PSM14,AAWSPGPC16},
and cyber-security~\cite{SSM15,SYLS18,RKVC18}.

In the early stage of machine learning, people pay more attention to the
basic theory and application research, although it is known in 2004 that machine learning models are often vulnerable to adversarial manipulation
of their input intended to cause misclassification~\cite{DDMSV04}.
In 2014, Szegedy et al. proposed the
concept of adversarial examples for the first time in deep neural network setting~\cite{SZSBEGF14}.
By adding a subtle perturbation to the input of the deep neural
network, it results in a misclassification.
Moreover, a relatively large fraction of adversarial examples can be used to attack
models that have different architectures and training data.
%Szegedy et al. formalized the computation of adversarial examples as the constrained optimization
%problem for which an approximate solution is sought using a
%box-constrained L-BFGS method.
%
Since these findings, a plethora of studies have shown that the state-of-the-art deep neural networks suffer
from the adversarial example attacks which can lead to severe consequences
when applied to real-world \mbox{applications}~\cite{LiV14,GSS15,NYC15,CW17a,CW17b,PMJFCS16,SBBR16,MFFF17,PCYJ17,KGB17,BRB18,XZLHLS18,ZDS18,KFS18,EEF0RXPKS18,AEIK18,CZSYH17,ilyas2017query,PMGJCS17,BHLS17,TTCLZYC18,CLCYZH18,WHK18,IEAL18}.
%How to measure robustness of deep neural networks becomes an urgent and important problem.
In the literature, there are mainly two types of complementary techniques:
testing based~\cite{SZSBEGF14,NYC15,PCYJ17,MFFF17,EEF0RXPKS18,MJZSXLCSLLZW18,AEIK18,PMGJCS17,IEAL18,ilyas2017query,WHK18,CLCYZH18,TTCLZYC18,BHLS17,KGB17,BRB18} and
verification based~\cite{KBDJK17,PT10,GMDTCV18,WHK18,GKPB18,SGMPV18,SGPV19}
methods for crafting adversarial examples.
According to the adversary's knowledge and capabilities,
these techniques also can be categorized into both white-box~\cite{SZSBEGF14,NYC15,PCYJ17,MFFF17,EEF0RXPKS18,MJZSXLCSLLZW18,AEIK18,KBDJK17,PCYJ17,PT10,GMDTCV18,GKPB18,SGMPV18} and
black-box~\cite{PMGJCS17,IEAL18,ilyas2017query,WHK18,CLCYZH18,TTCLZYC18,BHLS17,KGB17,BRB18,WHK18},
where white-box attacks require full white-box access to the target model,
which is not always feasible in practice.

%Prior explorations of adversarial examples largely focus on digital adversarial examples~\cite{GSS15,NYC15,MFFF17,WHK18,CLCYZH18,CW17b,GMDTCV18}.
%More recently, researchers started to study physical adversarial examples. Kurakin et al. showed that there is a big gap between adversarial examples in the digital world and in the physical world, which means the adversarial perturbations that generalize well in the digital world may not generalize to the physical world~\cite{KGB17}.
%
%Besides this gap,

However,  almost all existing adversarial example attacks target neural networks
rather than neural network based classifiers, while neural network based classifiers differ from
neural networks.
As a matter of fact, in image classification setting,
valid images in computer systems are stored in some format (e.g., png and jpeg) formed as a discrete integer  domain (e.g., $\{0,\cdots,255\}^m$),
but  will be normalized into some continuous real domain (e.g., $[0,1]^m$) for training and testing neural network models~\cite{GoodfellowBC16}. Therefore, a neural network based image classifier
consists of a pre-processor for normalization and a neural network model.
As a result, adversarial examples crafted by existing attacks against
neural networks are in the continuous real domain.
Such adversarial examples do fool the target neural network,
but once denormalized back into the discrete integer domain as valid images,
may become benign for the neural network based image classifier.
%Meanwhile, it is found out that there exists a gap between adversarial examples in the continuous (real) domain (e.g., $[0,1]^m$) and in the discrete (integer) domain (e.g., $\{0,\cdots,255\}^m$),
%although both of them are in the digital world.
%This gap comes from the fact that: valid images in discrete domain
%are usually normalized into a continuous domain (e.g., $[0,1]$), on which deep learning models are %trained and tested;
%whereas adversarial examples crafted in the continuous domain may become benign
%after denormalization post-processing which  denormalize adversarial examples back
%into the discrete integer domain.
This gap was initially considered by Goodfellow et al.~\cite{GSS15} and Papernot et al.~\cite{PMJFCS16},
and latter formally presented by Carlini and Wagner, called \emph{the discretization problem}~\cite{CW17b}.
Carlini and Wagner stated that ``\emph{This rounding will slightly degrade the quality of the
adversarial example}'' according to their experimental results on MNIST images.
Later on, this problem has received relatively little attention.
We believe, there lacks a comprehensive study on the impacts of the discretization problem:
e.g., which methods/tools may be affected,
to what extent does this problem affect the attack success rate and
can it be avoided or alleviated?

%However, it seems that the discretization problem was underestimated in the literature and has received relatively little attention.
%For instance, Carlini and Wanger~\cite{CW17b} stated that ``\emph{This rounding will slightly degrade the quality of the
%adversarial example}''.

%We called adversarial examples in the continuous domain, \emph{real} adversarial examples
%and adversarial examples in the discrete domain, \emph{integer} adversarial examples.

%Traditional attacks first normalize an input image from a discrete domain into a continuous domain,
%then craft adversarial examples in the continuous domain and finally denormalize (e.g., discretize) real adversarial examples back into the discrete domain.
%However, real adversarial examples may become benign after the denormalization post-processing, which is called the discretization problem in~\cite{CW17b}.
%This discretization problem was initially considered by Goodfellow et al.~\cite{GSS15} and Papernot et al.~\cite{PMJFCS16}.
%A follow-up paper~\cite{CW17b} formally mentioned this problem.
%However, it seems that the discretization problem was underestimated in the literature and has received relatively little attention.
%For instance, Carlini and Wanger~\cite{CW17b} stated that ``\emph{This rounding will slightly degrade the quality of the
%adversarial example}''.
%There lacks a study on the impacts of the discretization problem:
%e.g., to what extent do the discretization problem affect the attack success rate and
%can the discretization problem  be avoided or alleviated?

To understand the impacts of the discretization problem,
in this work, we report the first comprehensive study of existing works  for crafting adversarial examples in image classification domain
which has a plethora of studies. %\footnote{Learning algorithms on other problems using discretization may have similar problem. We leave it as future work.}.
In the rest of this work, adversarial examples in a continuous real domain will be called \emph{real} adversarial examples
and adversarial examples in a discrete integer domain will be called \emph{integer} adversarial examples.

We first discuss the difference between adversarial examples in the continuous domain and in the discrete domain,
%as well as distance metrics which are commonly used to approximate human's perception of visual difference.
%Then, we revise the definition of black-box attacks.
%We found that almost all the traditional black-box attacks that are assumed to have access to the normalization and denormalization between
%vector and discrete images, and many complete verification methods are only limited
%in continuous vector domain, which become incomplete
%in discrete integer domain.
%
 % Hence works that claimed are black-box attacks downgrade to grey-box.
%
%
%,
%and check whether these methods craft real or integer adversarial examples,
%whether there is gap between the real and integer adversarial examples,
%and whether the authors are aware of the discretization problem and take it into account.
%
Then, we theoretically analyze 35 representative methods for crafting adversarial examples.
We find that:
\begin{itemize}
\item Almost all of them craft real adversarial examples;
\item 29 methods are affected by the discretization problem;
\item 23 works do not provide hyper-parameters so that the discretization problem could not be easily and directly avoided.
\end{itemize}

%
%(3) 2 methods that are claimed as black-box downgrade to gray-box (i.e., the
%adversary has partial knowledge, i.e., normalization, of the classifier);
%and (4) 4 complete verification methods that are limited to a continuous domain only become incomplete
%in the discrete domain.
To understand the impacts of the discretization problem in practice,
we carry out an empirical evaluation of $20$ representative open source tools. We evaluate the gap between the attack success rates of crafted real adversarial examples and their corresponding integer adversarial examples. Our empirical study shows that:
\begin{itemize}
\item Most of the $20$ tools are affected by the discretization problem. In our experiments, there are 8 tools whose gap exceeds 50\%, 6 tools whose gap exceeds 70\%, and only 6 tools do not have any gaps.
\item Among the 14 tools that are affected by the discretization problem, only 1 tool (FGSM) can avoid the discretization problem by tuning input parameters,
3 tools can alleviate the discretization problem by tuning input parameters at the cost of attack efficiency or imperceptibility of adversarial examples,
and 10 tools can neither avoid nor alleviate the discretization problem by tuning input parameters.
\end{itemize}
%
%
%(1) perturbation step size may significantly affect the
%attack success rate due to the discretization problem;
%(2) many tools are affected by the discretization problem using the commonly
%adopted default input parameters (there are 10 tools whose
%gap exceeds 50\% and 7 tools whose gap exceeds 80\%)
%and (3) the discretization problem can be avoided or alleviated by tuning input parameters or using the greedy search algorithm of
%Carlini and Wagner~\cite{CW17b} for many white-box tools, but it seems difficult to alleviate for many tools  (e.g., DeepXplore, Bandits and verification tools).
\noindent\emph{Our study reveals that the discretization problem is far more serious than originally thought and
suggests to take it into account seriously
when crafting digital adversarial examples and measuring attack success rate.}

%A naive idea to avoid the discretization problem is to guarantee
%that no gap occurs when denormalizing the real adversarial examples back into the discrete domain.
%However, this requires access to the denormalization and the attack become grey-box,
%which is not always feasible in practice. A ``real" black-box attack should not have access to
%the model, as well as the normalization and denormalization.
%Existing black-box attacks that rely on the transferability property of adversarial examples could avoid this problem,
%but require a model that is similar to the target model and re-trained on a similar dataset.

\medskip
According to our comprehensive study, we found
there lacks an effective and efficient integer adversarial example attack
in black-box scenario.
As the second main contribution of this work,
we propose a black-box algorithm that directly crafts adversarial examples in discrete integer domains for both targeted and untargeted attacks.
Our method only requires access to the probability distribution of classes for each test input.
% We compute integer perturbations of images in the discrete integer domain,
% which always results invalid images.
We formalize the computation of integer adversarial examples as a black-box discrete optimization problem
constrained with a $\mathds{L}_\infty$ distance, where $\mathds{L}_\infty$ is defined in the discrete domain as well.
However, this discrete optimization problem cannot be solved using gradient-based methods, as
the model is non-continuous.
To solve this problem, we propose a novel classification model-based derivative-free discrete optimization method that
does not rely on the gradient of the objective function, but instead,
\emph{learns} from samples of the search space and \emph{refines} the search space into small sub-spaces.
It is suitable for optimizing functions that are non-differentiable,
with many local minima, or even unknown but only testable.

We demonstrate the effectiveness and efficiency of our method
on the MNIST dataset~\cite{MNIST98} using the LeNet-1 model~\cite{LBBH98};
and the ImageNet dataset~\cite{DDSLL009} using  Inception-v3~\cite{SVISW16} model.
Our method achieves close to 100\% attack success rates for both targeted
and untargeted attacks, comparable to the state-of-the-art white-box attacks:
FGSM~\cite{GSS15}, BIM~\cite{KGB17} and C\&W~\cite{CW17b},
and significantly outperforms representative black-box methods:
ZOO~\cite{CZSYH17},
AutoZOOM~\cite{TTCLZYC18},
NES-PGD~\cite{IEAL18},
Bandits~\cite{IEM2018PriorCB},
GenAttack~\cite{ASCZHS19},
substitute model based black-box attacks with
FGSM and C\&W methods,
FD and FD-PSO~\cite{BHLS18}.
In terms of query efficiency,
our attack is comparable to (or better than) the black-box attacks: NES-PGD, Bandits, AutoZOOM, and GenAttack, which are specially designed for query-limited scenarios.
%
%
%$>3$ times less than the query-limited attack~\cite{IEAL18} for each
%successful untargeted attack, and is $>4.8$ times less than it for each
%successful targeted attack.
Moreover, our method is able to break
the HGD defense \cite{LLDPH018}, which won the first place of
NIPS 2017 competition on defense against adversarial attacks, with
100\% success rate, and also achieves the so-far best success rate of white-box attacks in the online MNIST Adversarial Examples Challenge~\cite{mnistchallenge19}.

Our contributions in this paper include:
\begin{itemize}%[noitemsep,topsep=0pt,leftmargin=*]
  \item We report the first comprehensive study of existing works on the discretization problem,
  including 35 representative methods and 20 representative open source tools.

 \item Our study sheds light on the impacts of the discretization problem, which is useful to the community. % in highlighting the potential impacts of discretization.

%  theoretically analyze 35 representative methods for crafting adversarial examples, and find that
 % 29 methods are affected by the discretization problem.
% 2 black-box attacks downgrade to gray-box ones, and 4 complete methods become incomplete, due to the discretization problem.

%  \item We empirically evaluate 20 representative open source tools that can craft adversarial examples,
 % and find that 14 of them have a gap between the success rates of real and integer adversarial examples.
  %, and the gap of 8 tools exceeds 50\% using their default input parameters.

  \item We propose a black-box algorithm for crafting integer adversarial examples for targeted/untargeted attacks by designing a derivative-free discrete optimization method.
%  Our method can attack machine learning models that are even not differentiable.

  \item Our attack achieves close to 100\% attack success rate,  comparable to several recent popular white-box attacks, and outperforms several recent popular black-box tools
   (e.g., ZOO,
Bandits,
AutoZOOM, GenAttack and
NES-PGD) in terms of  integer adversarial examples.
%FGSM~\cite{GSS15} and C\&W~\cite{CW17b},
%and black-box based attacks:
%ZOO~\cite{CZSYH17}, DBA~\cite{BRB18},
%NES-PGD~\cite{IEAL18}, and also FGSM~\cite{GSS15} and C\&W~\cite{CW17b} with a substitute model.

 % \item We demonstrate that our attack can be used as a query-efficient black-box attack. It uses several hundreds of to thousands of query times less than  NES-PGD and Bandits, which are specially designed for query-limited scenarios.

\item Our attack is able to break
the HGD defense \cite{LLDPH018} with
100\% success rate, and also achieves the same result as the best white-box attack in MNIST Challenge~\cite{mnistchallenge19}.
\end{itemize}

\medskip
\noindent
{To the best of our knowledge, this is the first comprehensive study of
the impacts of the discretization problem on adversarial examples
and the first black-box attack that directly crafts adversarial examples in discrete integer domain.}

%, .

%% file: related_work.tex
\section{Related Work}
\label{sec:related}
Digital adversarial attacks in white-box scenario have been widely
studied in the literature, to cite a few \cite{SZSBEGF14,NYC15,PCYJ17,MFFF17,EEF0RXPKS18,MJZSXLCSLLZW18,AEIK18,KBDJK17,PCYJ17,PT10,GMDTCV18,GKPB18,SGMPV18}.
In white-box scenario, the adversary has access to details (e.g., architecture, parameters, training dataset) of the system under attack.
This setting is clearly impractical in real-world cases,
when the adversary cannot get access to the details.
Therefore, in this work, we propose black-box adversarial attacks.
In the rest of this section,
we mainly discuss existing works on black-box adversarial attacks .

\subsection{Digital Adversarial Attack}
We classify existing attack methods
along three dimensions: substitute model, gradient estimation and heuristic search.

 \medskip
\noindent{\bf{Substitute Model.}}
Papernot et al.~\cite{PMGJCS17} proposed the first black-box method by leveraging transferability
property of adversarial examples.
It first trains a local substitute model with a synthetic dataset and then crafts adversarial examples from the local substitute model.
\cite{PMG16} generalized this idea to attack other machine learning classifiers.
However, transferability is not always reliable, other methods such as gradient estimation are explored as alternatives to substitute networks.

 \medskip
\noindent{\bf{Gradient Estimation.}}
Gradient plays an important role in white-box adversarial attacks. Therefore, estimating the gradient to guide the search of adversarial examples is a popular research direction in black-box adversarial attacks.
Narodytska and Kasiviswanathan~\cite{NK17} proposed
a greedy local search based method to construct
numerical approximation to the network gradient,
which is then used to construct a small set of pixels in
an image to perturb.
Chen et al.~\cite{CZSYH17} proposed a black-box attack method (named ZOO) with zeroth order optimization. Following ZOO, Tu et al.~\cite{TTCLZYC18} proposed an autoencoder-based method (named AutoZOOM) to improve query efficiency. 
Similarly, Bhagoji et al.~\cite{BHLS18} proposed a class of black-box attacks (called FD) that approximate FGSM and BIM via gradient estimation.
Independently, Ilyas et al.~\cite{IEAL18} proposed an alternative gradient estimation method by leveraging natural evolution strategy (NES)~\cite{SHCS17,WSGSPS14}
and employing white-box PGD attack with estimated gradient (named NES-PGD).
Based on NES-PGS, Ilyas et al.~\cite{IEM2018PriorCB} proposed a bandit optimization-based method aimed at enhancing query efficiency. Recently, Zhao et al.~\cite{zoadmm19} proposed a method to leverage an alternating direction method of multipliers (ADMM) algorithm for gradient estimation.

 \medskip
\noindent{\bf{Heuristic Search.}}
Instead of gradient estimation,  heuristic search-based derivative-free optimization (DFO) methods have been proposed.
%such as EvadeML~\cite{xu2016automatically} using genetic programming,
%accessorize-to-a-crime~\cite{SBBR16} using particle swarm optimization,
%NES-PGD~\cite{IEAL18} using natural evolution strategies,
%ZOO~\cite{CZSYH17} and AutoZOOM~\cite{TTCLZYC18} using zeroth order
%optimization,
%and Bandits~\cite{IEM2018PriorCB} using bandit optimization.
%These works use heuristic search or gradient estimation to solve derivative-free problems.
%derivative-free optimization methods for either gradient estimation or directly finding solutions without refining search space.
Hosseini et al.~\cite{HXP17} proposed a method
by iteratively adding Gaussian noise.
Liu et al.~\cite{LCLS17} proposed ensemble-based approaches to generating transferable adversarial examples.
Brendel et al.~\cite{BRB18} proposed a decision-based attack (named DBA) with label-only setting, which starts from the target image, moves a small step to raw image every time and checks the perturbation cross the decision boundary or not.
Su et al.~\cite{SU17} proposed a black-box attack for generating one-pixel adversarial images based on differential evolution.
Bhagoji et al.~\cite{BHLS18} also proposed a particle swarm optimization (PSO) based DFO method, named FD-PSO. PSO previously was used to find adversarial examples to fool face recognition
systems~\cite{SBBR16}. In a concurrent work, Alzantot et al.~\cite{ASCZHS19} proposed a genetic algorithm based DFO method (named GenAttack) for generating adversarial images.
Genetic algorithm was previously used to find adversarial examples to fool PDF malware classifiers in EvadeML~\cite{xu2016automatically}.
Co et al.~\cite{co2019procedural} proposed
a method for generating universal adversarial perturbations (UAPs) in the black-box attack scenario
by leveraging Bayesian optimization,
it is a new interesting area to generate procedural noise perturbations.

 %In Section \ref{sec:expr}, we conduct throughly experimental studies of our method and also compare our method with state-of-the-art tools from all the above classes. We can see that our tool achieves significantly higher success rate in terms of the integer adversarial examples than all the other tools, with comparable query times.

 \medskip
\noindent{\bf{Comparison.}} Our method
does not rely on substitute model or gradient estimation.
Different from the above heuristic search based methods, we present a classification model-based DFO method, to distinguish ``good'' samples with ``bad'' samples. By learning from the evaluation of the samples,  our algorithm iteratively refines
large search space into small-subspaces, finally converges to the best solution.
To the best of our knowledge, our method is the first one which iteratively refines
large search space into small-subspaces during searching adversarial examples. 
Experimental results show that our method
achieves significantly higher success rate in terms of the integer adversarial examples than the state-of-the-art tools from all the above classes, with comparable query times (cf. Section \ref{sec:expr}).

Although, some of these works (e.g.,~\cite{HXP17,LCLS17,BRB18}) for crafting digital adversarial samples add noises onto integer images and clip the value of each pixel
into the range of 0 and 255,  the noise added to each coordinate could be real numbers and the value of each coordinate is not clipped in the discrete integer domain $\{0,\cdots, 255\}$.
Therefore, their methods may craft many useless invalid integer images, reducing efficiency. 
While our method directly crafts adversarial samples the discrete integer domain, hence avoids 
to craft useless invalid integer images.

\subsection{Physical Adversarial Attack}
Thanks to the success of adversarial example attacks in the digital domain,
recently, researchers started to study the feasibility
of adversarial examples in the physical world.
We now discuss recent efforts on physical adversarial examples.

Kurakin showed that printed adversarial examples crafted in the
digital domain can be misclassified when viewed through a smartphone camera~\cite{KGB17}.
Follow-up works proposed methods to improve robustness of physical adversarial examples
by synthesizing the digital images
to simulate the effect of rotation, brightness and scaling, and digital-to-physical transformation~\cite{LSF17,EEF0RXPKS18,AEIK18,JMLHW19},
or manually taking physical photos from different viewpoints
and distances~\cite{EEFLSKRPT17,EEF0RXPKS18},
or adding scene-independent patch~\cite{BMRAG17}.
Furthermore, adversarial example attacks have been applied on road sign images~\cite{LuSFF17,SBMCM18}, face recognition systems~\cite{SBBR16} and object detectors~\cite{CCMC18,SEEF0RTPK18}.
Physical adversarial examples that are printed or showed by devices
will not be affected by the discretization
problem.

Although, these works demonstrated that physical adversarial examples are possible, and
integer adversarial images may be damaged by image transformations (e.g., photo, brightness, contrast, and etc.) in the physical world~\cite{KGB17},
it is still very useful to generate effective integer adversarial images.
\begin{itemize}
\item First, it can be used in many practical scenarios, e.g., attacking the online image classification systems.
\item Second, an attacker who cannot fool a classifier successfully in the digital domain will also struggle to do so physically in practice~\cite{SBBR16}.
\item Third, it usually requires relatively expensive manual efforts to directly
craft physical adversarial examples. On the other hand, robust digital adversarial examples can survive in physical world~\cite{JMLHW19}.
\end{itemize}

It is interesting to study the impacts of the discretization problem on the difficulty of finding physical adversarial examples.
To apply our classification model-based derivative free optimization method on physical attack is also an interesting topic.
We leave these topics to future work.

\subsection{Other Attacks}
Adversarial example attacks against other machine learning based classifiers also have been exhibited,
such as malicious PDF files~\cite{maiorca2013looking,SL14,xu2016automatically},
malware~\cite{grosse2017adversarial,demontis2017yes}, malicious websites~\cite{xu2014evasion}, spam emails~\cite{lowd2005good},
and speech recognition~\cite{YuanCZLL0ZH0G18,CarliniMVZSSWZ16}.
Since each type of machine learning based classifiers has unique characteristics, in general, these existing
attacks are orthogonal to our work.

%Jan et al.~\cite{JMLHW19} proposed an image-to-image translation network to simulate the digital-to-physical transformation process for generating robust adversarial examples.

%RACOS~\cite{yu2016derivative}, with lots of domain-specific optimizations to improve  efficiency and scalability, for finding solutions. The main advantage of RACOS is that it refines search space during finding solutions.

%% file: background.tex
\section{Background}
\label{sec:back}
In this section, we introduce deep learning based image classifications, adversarial attacks and distance metrics.
For convenient reference, we summarize the notations in Table~\ref{tab:notations}.

%Real number arithmetic is implemented using floating-point arithmetic in computers which may suffer from the rounding errors.
%In this work, we assume that real number arithmetic is sound~\cite{SGMPV18,SGPV19}.

\begin{table}[t]\small
\caption{Notations used in this paper}
\label{tab:notations}
\begin{tabular}{l|l} %{p{2.8cm}|p{5cm}}
  \toprule
  {\bf Notation} & {\bf Description} \\ \midrule

   $w$, $h$, ${ch}$ & \tabincell{l}{width, height, and number of channels of an image} \\ \midrule
   $P$ &  the set of coordinates $w\times h \times {ch}$ \\ \midrule
  $\mathds{V}$ & \tabincell{l}{{\bf continuous (real)} domain of {\bf real} images $\vec{v}$,  e.g., $\mathds{R}_{[0,1]}^{w\times h \times {ch}}$} \\ \midrule
 $\mathds{D}$ &  \tabincell{l}{{\bf discrete (integer)} domain of {\bf integer} images  $\vec{d}$,  e.g., $\mathds{N}_{[0,255]}^{w\times h \times {ch}}$} \\ \midrule
$\vec{v},~\vec{v}^{\adv}\in \mathds{V}$ & continuous \textbf{real} (adversarial) image\\ \midrule
$\vec{d},~\vec{d}^{\adv}\in \mathds{D}$ & discrete \textbf{integer} (adversarial) image\\ \midrule
  $\vec{v}[p]$ & {entity at coordinate $p$ of a {\bf real} image $\vec{v}$}\\ \midrule
 $\vec{d}[p]$ & entity at coordinate $p$ of an {\bf integer} image $\vec{d}\in \mathds{D}$ \\ \midrule
$\mathds{T}:\mathds{D}\rightarrow\mathds{V}$ & \tabincell{l}{normalizer that transforms an \textbf{integer} image  into a  \textbf{real} image \\in the continuous domain $\mathds{V}$} \\ \midrule %an injective but non-surjective function, called
 $\mathds{T}^{-1}:\mathds{V}\rightarrow\mathds{D}$ &  \tabincell{l}{denormalizer that transforms a \textbf{real} image back into an  \textbf{integer} image  such that \\ for all $\vec{d}\in \mathds{D}, \ \mathds{T}^{-1}(\mathds{T}(\vec{d}))=\vec{d}$} \\ \midrule
$\mathds{C}_t$ & \tabincell{l}{set of mutually exclusive classes for the task $t$}  \\ \bottomrule
\end{tabular}% \vspace{-3mm}
\end{table}

\subsection{Deep Learning based Image Classification}
\label{sec:dnnclass}
\begin{figure}[h]
  \centering
  \includegraphics[width=0.7\textwidth]{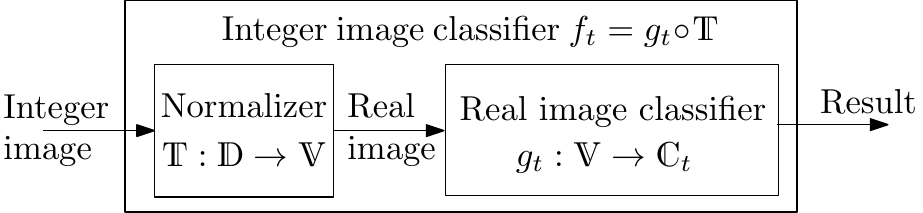}
  \caption{Overview of machine learning based image classifiers.}\label{fig:classifier}
\end{figure}

Valid images are represented as integer images in computer systems.
To train a practical image classifier $f_t:\mathds{D}\rightarrow \mathds{C}_t$,
valid images should first be normalized so that their pixels all lie in the same reasonable range,
as integer images come in a form that is difficult for many deep learning architectures to
represent~\cite{GoodfellowBC16}.
Therefore, as shown in Figure~\ref{fig:classifier}, the classifier $f_t$
is constructed by training an image classifier $g_t: \mathds{V}\rightarrow \mathds{C}_t$ in continuous (real) domain aided by a normalizer $\mathds{T}:\mathds{D}\rightarrow\mathds{V}$,
which leads to the classifier $f_t=g_t\circ \mathds{T}$.

%Deep learning methods are used to train a classifier $\widehat{g}_t: \mathds{V}\rightarrow \mathds{C}_t$ as an approximation of the expected one $g_t$.
%Consider a supervised learning task $t$, let $(d,c_d)_{d\in D}$ be a finite set of labelled data, where $c_d\in \mathds{C}_t$ denotes the class of the integer image $d\in D$.
%The labelled dataset $(d,c_d)_{d\in D}$ is normalized into the dataset $(v,c_v)_{v\in V}$  by the normalizer $\mathds{T}$,
%where $V=\{\mathds{T}(d)\mid d\in D\}$ and $c_v=c_d$ if $v=\mathds{T}(d)$ for $d\in D$ and $v\in V$.
%A deep neural network is trained and tuned based on the dataset $(v,c_v)_{v\in V}$,  resulting
%in the classifier $\widehat{g}_t$. Finally, the classifier $f_t$ is approximated by the function
%$\widehat{f}_t:=\widehat{g}_t\circ \mathds{T}$.
%Figure~\ref{fig:models} depicts the relationship between these notations for convenient.
%Remark that normalizers may differ in tools, neural network models and dataset (cf. Appendix~\ref{sec:discnorm}).

\subsection{Adversarial Attacks}
In this work, we consider adversarial attacks using digital adversarial examples instead of
physical adversarial examples.
We categorize digital adversarial examples
into \emph{real} and \emph{integer} ones
according to their domains $\mathds{V}$ and $\mathds{D}$.

\medskip
\noindent{\bf Real and integer adversarial examples.}
A \emph{real adversarial example} crafted from a real image $\vec{v}\in \mathds{V}$
is an image $\vec{v}^{\adv}\in \mathds{V}$ such that
the real image classifier ${g}_{t}$ misclassifies $\vec{v}^{\adv}$,
i.e., \[{g}_{t}(\vec{v})\neq {g}_{t}(\vec{v}^{\adv}).\]
Likewise, an \emph{integer adversarial example} crafted from an integer image $\vec{d}\in \mathds{D}$
is an image  $\vec{d}^{\adv}\in \mathds{D}$
such that the integer image classifier ${f}_{t}$ misclassifies   $\vec{d}^{\adv}$,
i.e., \[{f}_{t}(\vec{d})\neq {f}_{t}(\vec{d}^{\adv}).\]

\medskip
\noindent{\bf Untargeted and targeted attacks.}
In the literature, there are two types of adversarial attacks: targeted and untargeted attacks.
Untargeted attack aims at crafting an adversarial example that misleads the system being attacked, i.e.,
${g}_{t}(\vec{v})\neq {g}_{t}(\vec{v}^{\adv})$ for real adversarial examples
and ${f}_{t}(\vec{d})\neq {f}_{t}(\vec{d}^{\adv})$ for integer adversarial examples.
A more powerful but difficult attack,  targeted attack, aims at crafting an adversarial example such that
the system classifies the adversarial example as the given class $c$,
i.e., ${g}_{t}(\vec{v}^{\adv})=c$ for real adversarial examples
and ${f}_{t}(\vec{d}^{\adv})=c$ for integer adversarial examples.
It is easy to see that targeted attack can be used to launch untargeted attack
by choosing an arbitrary target class.

%
%Given a classifier $(\widehat{g}_{t},\mathds{T})$ and an integer image $\vec{d}\in \mathds{D}$,
%an \emph{integer adversarial example} for $\vec{d}$ is an integer image $\vec{d}^{\adv}\in \mathds{D}$
%such that
%\begin{center}
%$\widehat{g}_{t}(\mathds{T}(\vec{d}))\neq \widehat{g}_{t}(\mathds{T}(\vec{d}^{\adv}))$.
%\end{center}
%
%Given a classifier $\widehat{g}_{t}$ and a (continuous) real image $\vec{v}\in \mathds{V}$ that may be obtained from some (discrete) integer image $\vec{d}\in \mathds{D}$ via the normalizer $\mathds{T}$,
%an \emph{real adversarial example} for $\vec{v}$ and $\widehat{g}_{t}$ is a real image $\vec{v}^{\adv}\in \mathds{V}$
%such that
%$\widehat{g}_{t}(\vec{v})\neq \widehat{g}_{t}(\vec{v}^{\adv})$,
% i.e., the classifier $\widehat{g}_{t}$ misclassifies the example $\vec{v}^{\adv}$.
%
%An \emph{untargeted attack} against the classifier $\widehat{g}_{t}$ is to craft a real adversarial example $\vec{v}^{\adv}$
%for a given real image $\vec{v}\in \mathds{V}$.
%A more powerful attack, \emph{targeted attack}, for a given target class $c$
%is to craft a real adversarial example $\vec{v}^{\adv}$ such that
%$\widehat{g}_{t}(\vec{v}^{\adv})=c$.
%%It is often expected that the real adversarial example $\vec{v}^{\adv}$ and the original one
%%$\vec{v}$ are visually indistinguishable by humans.
%

\medskip
\noindent{\bf White-box and black-box scenarios.}
Targeted and untargeted attacks have been studied in
both white-box and black-box scenarios, according to the knowledge of the target system.
In white-box scenario, the adversary has access to details (e.g., architecture, parameters and training dataset) of the system under attack.
This setting is clearly impractical in real-world cases,
when the adversary cannot get access to the details.
In a more realistic black-box scenario, it is usually assumed
that the adversary can only query the system and obtain
confidences or probabilities of classes for each input by limited queries.

In black-box scenario, we emphasize that the adversary has no access to the normalization
of the target classification system, otherwise the attack would be a gray-box one.
It is also non-trivial to infer the normalization by the adversary in black-box scenario due to the diversity of normalization.
Indeed, there is no standard normalization in literature
and they may differ in tools, neural network models and datasets.
For instance,  let $i$ denote the integer value of a coordinate,
\begin{itemize}
  \item the Inception-v3 model on ImageNet dataset in ZOO~\cite{CZSYH17}  uses the normalization:
  \[v_1=(i/255 -0.5);\]
\item  the Inception-v3 model on ImageNet dataset in Keras~\cite{keras19} uses the normalization:
\[v_2=((2\times i)/255 -1);\]
\item the VGG and ResNet models on ImageNet dataset in Keras~\cite{keras19} use the normalization:
\[v_3=(i-mean),\]
where $mean$ denotes the mean value of images in training dataset.
\end{itemize}

To the best of our knowledge, there is no work on inferring normalization of classifiers.
More details refer to Appendix~\ref{sec:discnorm}.

\subsection{Distance Metrics}
The distortion of adversarial examples should be visually indistinguishable from their normal counterparts
by humans. However, it is hard to model human perception, hence several distance metrics were proposed to approximate human's perception of visual difference.
In the literature, there are four common distance metrics ${\bf L}_0$, ${\bf L}_1$, ${\bf L}_2$ and ${\bf L}_\infty$ which are defined over samples in
some continuous domain $\mathds{V}$.
All of them are ${\bf L}_n$ norm defined as
\[\|\vec{v}-\vec{v}^{\adv}\|_n= \left(\sum_{p\in P} \left|\vec{v}[p]-\vec{v}^{\adv}[p]\right|^n \right)^{\frac{1}{n}},\]
where $\vec{v},\vec{v}^{\adv}\in \mathds{V}$.
In more detail, ${\bf L}_0$ counts the number of different coordinates, i.e., $\sum_{p\in P} (\vec{v}[p]\neq \vec{v}^{\adv}[p])$;
${\bf L}_1$ denotes the sum of absolute differences of each coordinate value, i.e., $\sum_{p\in P} (\left|\vec{v}[p]- \vec{v}^{\adv}[p]\right|)$;
${\bf L}_2$ denotes Euclidean or root-mean-square distance;
and ${\bf L}_\infty$ measures the largest change introduced. Remark that
\[\lim_{n\rightarrow \infty}\|\vec{v}-\vec{v}^{\adv}\|_n=\max\{\left|\vec{v}[p]-\vec{v}^{\adv}[p]\right| \mid p\in P \}.\]

%We will denote by $\|\vec{v}-\vec{v}^{\adv}\|_{\bf L}$ the distance
%between $\vec{v}$ and $\vec{v}^{\adv}$ based on the distance metric ${\bf L}$.

However, %for a given metric ${\bf L}_n$, $\|\vec{v}^{\adv}-\mathds{T}(\vec{d})\|_n$ may differ from
%$\|\mathds{T}(\mathds{T}^{-1}(\vec{v}^{\adv}))-\mathds{T}(\vec{d})\|_n$.
it seems not reasonable to approximate human's perception of visual difference using distance metrics defined between real images.
Instead, it is much better to measure the distance between integer images.
For this purpose, we revise distance metrics and introduce $\mathds{L}_p$ norm which is defined between integer images.
Formally, $\mathds{L}_n$ is defined as follows:
%\begin{center} $\|\vec{d}-\vec{d}^{\adv}\|_n= \left(\sum_{p\in P} |\frac{\vec{d}[p]-\vec{d}^{\adv}[p]}{255}|^n \right)^{-n}$,\end{center}
\[\|\vec{d}-\vec{d}^{\adv}\|_n= \left(\sum_{p\in P} \left|\vec{d}[p]-\vec{d}^{\adv}[p]\right|^n \right)^{\frac{1}{n}},\]
where $\vec{d},\vec{d}^{\adv}\in \mathds{D}$.
Accordingly, we define: $\mathds{L}_0=\|\vec{d}-\vec{d}^{\adv}\|_0$, $\mathds{L}_1=\|\vec{d}-\vec{d}^{\adv}\|_1$,
$\mathds{L}_2=\|\vec{d}-\vec{d}^{\adv}\|_2$ and $\mathds{L}_\infty=\|\vec{d}-\vec{d}^{\adv}\|_\infty$.
Obviously, $\mathds{L}_n$ differs from ${\bf L}_n$ for any $n$.

%% file: discretization.tex
\section{The Discretization Problem}\label{sec:DPprobleminWild}
Recall that we categorize digital adversarial examples into real and integer ones according to their domains.
There is a gap between adversarial examples in continuous
and in discrete domains.
In this section, we first
formalize the gap as the discretization problem
and then report the comprehensively study of the impacts
of the discretization problem.

\subsection{Formulation of The Discretization Problem}
Recall that a practical image classification system $f_t$ is an integer
image classifier that consists of both
the real image classifier ${g}_{t}$
and the normalizer $\mathds{T}$.
Therefore, to attack the system $f_t$ using a real adversarial image $\vec{v}^{\adv}\in \mathds{V}$
that is crafted by querying ${g}_{t}$,
it is necessary to denormalize the real image $\vec{v}^{\adv}$ back into
a valid image (i.e, an integer image) $\vec{d}^{\adv}\in \mathds{D}$, so that it can be fed to the target system $f_t$.
To denormalize $\vec{v}^{\adv}$,
a denormalizer $\mathds{T}^{-1}$ should be implemented according to the knowledge of the normalizer $\mathds{T}$  such that  for any integer image $\vec{d}\in \mathds{D}$,
$\mathds{T}^{-1}(\mathds{T}(\vec{d}))=\vec{d}$.

However, after applying the denormalization,
$\vec{d}^{\adv}$ may be classified
as a class that differs from the one of  $\vec{v}^{\adv}$,
i.e.,
\[f_t(\vec{d}^{\adv})=f_t(\mathds{T}^{-1}(\vec{v}^{\adv})) ={g}_{t}(\mathds{T}(\mathds{T}^{-1}(\vec{v}^{\adv})))\neq g_t(\vec{v}^{\adv}).\]
This is so-called \emph{the discretization problem}~\footnote{The term ``discretization'' comes from~Carlini and Wagner\cite{CW17b} which expresses the rounding problem from real numbers to integer numbers.
Our definition is more general than theirs.},
which comes from the non-equivalent transformation between continuous real
and discrete integer domains, i.e., $\mathds{T}(\mathds{T}^{-1}(\vec{v}^{\adv}))\neq \vec{v}^{\adv}$,
resulting in
\[g_t(\mathds{T}(\mathds{T}^{-1}(\vec{v}^{\adv})))\neq g_t(\vec{v}^{\adv}).\]
In the rest of this work, the maximum error when transforming a real adversarial image back into the
discrete domain is called \emph{discretization error}.

The discretization problem may result in failure of untargeted and targeted attacks, i.e.,
\[f_t(\mathds{T}^{-1}(\vec{v}^{\adv}))=f_t(\vec{d}) \mbox{ or } f_t(\mathds{T}^{-1}(\vec{v}^{\adv}))\neq c.\] 
where $\vec{v}^{\adv}$ denotes a real adversarial image crafted from $\mathds{T}(\vec{d})$
and  $c$ denotes the target class.

As stated by Carlini and Wanger~\cite{CW17b},
the discretization problem slightly degrades the quality of the
adversarial example.
However, there lacks a comprehensive study of
the impacts of the discretization problem.
In the rest of this section, we report the first
comprehensive study including theoretically analysis of 35 representative methods and empirically study of 20 representative open source tools,
in an attempt to understand the impacts of the discretization problem. % in the wild.

\subsection{Theoretical Study}\label{sec:theory_study}
%Let us fix a classifier $\widehat{g}_t$ for an image classification task $t$,
%suppose $\vec{v}\in \mathds{V}$ is the real image from the integer image $\vec{d}\in \mathds{D}$ using a normalizer $\mathds{T}$ and the target class is $c$ for targeted attack
%We assume that $\widehat{g}_t(\vec{v})=c_{\vec{v}}\neq c$ and
%the loss function is denoted by $\loss:\mathds{V}\times \mathds{C}_t\rightarrow\mathds{R}$.

\begin{table}
\centering \small
\caption{Summary of theoretical study results, where
\emph{(un)targeted} column shows the type of attack, once a method could launch targeted attack, we mark it as targeted, as targeted is more powerful than untargeted attack;
\emph{Domain} column shows the domain of images;
%\emph{Distance} column shows the considered distance metrics;
\emph{Considered} column shows whether the method considered the discretization problem;
% affected by the ;
\emph{B2G} column shows whether black-box downgrades to gray-box;
\emph{Avoidable} column shows whether the discretization problem could be (almost) avoided;
\emph{Complete} column shows whether the method is complete,
$\to$ meaning complete method becomes incomplete due to the discretization problem}
\label{tab:TheStuofDPI}

\begin{tabular}{|c|c|c|c|c|c|c|c|}
\toprule
& & {\bf Reference} & {\bf (Un)targeted}  & {\bf Domain} & {\bf Considered} & {\bf B2G} & {\bf Avoidable}
\\ \hline
  \multirow{24}{*}{\rotatebox{90}{\bf Testing-based methods}} &  \multirow{16}{*}{\rotatebox{90}{\bf White-box}}
  &  L-BFGS~\cite{SZSBEGF14}   & Targeted     & Continuous  & \xmark & -  & \xmark   \\     \cline{3-8}
  &  & FGSM~\cite{GSS15}       & Untargeted   & Continuous  & \cmark & -  & \cmark \\    \cline{3-8}
  &  & BIM(ILLC)~\cite{KGB17}  & Targeted     & Discrete    & - & -  & - \\    \cline{3-8} %Both
  &  & PGD~\cite{MyMSTV17}     & Untargeted   & Continuous    &\xmark & -  & \xmark \\    \cline{3-8}
  % & & ILLC~\cite{KGB17}      & Targeted     & Discrete    &- & -  & -   \\    \cline{3-8}
  &  & MBIM~\cite{DLPSZHL18}   & Targeted   & Continuous   		&  \xmark &  -  & \cmark   \\   \cline{3-8} %Both
  &  & JSMA~\cite{PMJFCS16}    & Targeted   & Continuous   &  \xmark &  -  & \cmark \\    \cline{3-8}
  &  & C\&W~\cite{CW17b}       & Targeted   & Continuous       &  \cmark &  - & \xmark   \\   \cline{3-8} %Both
  &  & OptMargin~\cite{HLS18}  & Untargeted   & Continuous   & \xmark &  -  & \xmark  \\ \cline{3-8}
  &  & EAD ~\cite{CSZYH18}     & Targeted  & Continuous    	& \xmark &  -  & \xmark  \\  \cline{3-8} % Both
  %  \multirow{5}{*}{\rotatebox{90}{Non-gradient}}
  &  & DeepFool~\cite{MFF16}   & Untargeted      & Continuous & \xmark &  -  & \xmark  \\  \cline{3-8}
  &  & UAP~\cite{MFFF17}       & Untargeted       & Continuous  & \xmark &  -  & \xmark  \\ \cline{3-8}
  &  & DeepXplore~\cite{PCYJ17} & Untargeted  & Continuous   & \xmark & - & \xmark  \\  \cline{3-8}
  &  & DeepCover~\cite{SHK18}   & Untargeted & Continuous   & \xmark & - & \xmark  \\  \cline{3-8}
  &  & DeepGauge~\cite{MJZSXLCSLLZW18} & Untargeted  & Continuous   & \xmark & - & -  \\  \cline{3-8}
  &  & DeepConcolic~\cite{SWRHKK18}    & Untargeted & Continuous  & \cmark & - & \xmark  \\  \cline{2-8}  \cline{2-8}
  &  \multirow{9}{*}{\rotatebox{90}{\bf  Black-box}}
  &    SModel~\cite{PMGJCS17} & Targeted   & Continuous  & \xmark & \xmark  & -  \\ \cline{3-8} % Both
  &  & PMG~\cite{PMG16}       & Untargeted    & Continuous  & \xmark & \xmark  & -  \\ \cline{3-8}
  &  & One-pixel~\cite{SU17}  & Targeted   & Continuous  	&\xmark & \xmark  & \xmark \\ \cline{3-8} % Targeted
 % &  &SPA~\cite{NK17}  & Both    & Continuous & - &\cmark & \cmark  & \xmark \\ \cline{3-8}
  &  & ZOO~\cite{CZSYH17}     & Targeted     & Continuous  & \xmark & \cmark  & \xmark \\   \cline{3-8}
  &  & FD~\cite{BHLS18}       & Targeted     & Continuous  	& \xmark & \xmark  & - \\   \cline{3-8} % Both
  &  & NES-PGD~\cite{IEAL18}  & Targeted     & Continuous  & \xmark & \xmark  & \xmark \\   \cline{3-8}
  &  & DBA~\cite{BRB18}       & Targeted     & Continuous  & \xmark & \xmark  & \xmark \\  \cline{3-8} % Both
  &  & Bandits~\cite{IEM2018PriorCB}  & Untargeted   & Continuous  & \xmark & \xmark  & \xmark \\   \cline{3-8}
  &  & AutoZOOM~\cite{TTCLZYC18}  & Targeted   & Continuous  & \xmark & \cmark  & \xmark \\   \cline{3-8} % Both
  &  & GenAttack~\cite{ASCZHS19}  & Targeted   & Continuous  &  \xmark & \cmark  & \xmark \\
\hline \hline
& & {\bf Reference} & {\bf Complete}  & {\bf Domain} & {\bf Considered} & {\bf B2G} & {\bf Avoidable}\\ \hline
\multirow{10}{*}{\rotatebox{90}{\bf Verification methods}}& \multirow{9}{*}{\rotatebox{90}{\bf White-box}}
& BILVNC~\cite{BILVNC16}    & \cmark $\to$ \xmark & Continuous    & \xmark & - & \xmark \\  \cline{3-8}
& & DLV~\cite{HKWW17}       & \xmark 			& Continuous      & \cmark & - & \cmark \\  \cline{3-8}
& & Planet~\cite{Ehl17}     & \cmark $\to$ \xmark  & Continuous   & \xmark & - & \xmark \\  \cline{3-8}
& & MIPVerify~\cite{TXT18}  & \cmark $\to$ \xmark  & Continuous   & \xmark & - & \xmark \\  \cline{3-8}
& & DeepZ~\cite{SGMPV18}    & \xmark 			& Continuous      & \xmark & - & \xmark \\  \cline{3-8}
& & DeepPoly~\cite{SGPV19}  & \xmark 			& Continuous      & \xmark & - & \xmark \\  \cline{3-8}
& & DeepGo~\cite{RHK18}     & \xmark 			& Continuous      & \xmark & - & \xmark \\  \cline{3-8}
& & ReluVal~\cite{WPWYJ18}  & \cmark $\to$ \xmark  & Continuous   & \xmark & - & \xmark \\  \cline{3-8}
& & DSGMK~\cite{DSGMK18}    & \cmark $\to$ \xmark  & Continuous   & \xmark & - & \xmark\\  \cline{2-8}
& {\rotatebox{90}{\bf B}}
& SafeCV~\cite{WHK18}       & \xmark 			& Continuous      & \cmark & \cmark & \cmark  \\ \bottomrule

\end{tabular}
\end{table}

We theoretically analyze 35 existing works including
25 testing methods (15 white-box and 10 black-box)  and 10 verification methods (9 white-box and 1 black-box),
to determine:
1) whether they generate adversarial examples in discrete or continuous domain?
2) if they use some continuous domain, do they consider the discretization problem and how do they deal with?
and 3) if they do not consider, could the discretization problem be avoided by tuning input parameters?
The summary of results is given in Table~\ref{tab:TheStuofDPI} according to raw papers (primarily) and source code.

\medskip
\noindent
{\bf Discrete or continuous.}
After examining the domain of all the 35 works, we found
only BIM defines the adversarial example
searching problem in discrete domains and uses the integer perturbation step sizes.
While the other 34 works craft adversarial examples in continuous domains, hence they may be affected by the discretization problem.

\medskip
\noindent
{\bf Considered or not.}
Among 34 works that craft adversarial example in continuous domains,
we found only five works (i.e., FGSM, C\&W, DeepConcolic, DLV and SafeCV)
do consider the discretization problem,
while the other 29 works do not, indicating that 29 out of 35 works are affected by the discretization problem.

Specifically, FGSM uses perturbation step sizes that correspond to the magnitude of the
smallest bit of an image so that the transformation between
continuous and discrete domains are almost equivalent, i.e.,  the discretization errors are nearly zero.
DLV verifies classifiers by means of discretization such
that the crafted real adversarial examples are still adversarial after denormalization.
SafeCV limits the perturbation of each pixel to the minimum or maximum values of coordinates.
Therefore, the discretization problem in FGSM, DLV and SafeCV are (almost) avoided.

In contrast, C\&W and DeepConcolic perform denormalization post-processing before checking crafted real images,
and C\&W also proposes a greedy algorithm
that searches integer adversarial examples on a lattice defined
by the discrete solutions by changing one pixel value at a time.
However, discrete solutions are computed by rounding real numbers of coordinates in real adversarial examples
to the nearest integers.
Therefore, DeepConcolic and C\&W either evade or alleviate the discretization problem,
% both approaches are able to alleviate,
but they \emph{cannot} essentially avoid it in theory,
as they may craft many useless real adversarial examples.

\medskip
\noindent
{\bf Avoidable or not.}  We further conduct an in-depth analysis
of 29 works that craft adversarial
example in continuous domains, but do not consider the discretization problem. We investigate whether
the discretization problem in these works can be easily and directly avoided by
tuning hyper-parameters.
We found that only MBIM and JSMA
could control the perturbation step sizes directly by hyper-parameters
% provide hyper-parameters that can directly control the perturbation step sizes
so that the discretization problem could be (almost) avoided
by choosing proper perturbation step sizes.

In contrast, 23 out of 29 works do not provide such hyper-parameters
so that the discretization problem could not be easily and directly avoided.
This is because that
\begin{itemize}
  \item PGD, DeepXplore, One-pixel, NES-PGD, DBA, Bandits and GenAttack introduce random perturbation step size or random noise,  %intending to strengthen their attack ability or
%simulate physical environment, but
making perturbation step size uncontrollable;
  \item L-BFGS, OptMargin, EAD, DeepFool, DeepCover, UAP,  ZOO and AutoZOOM directly craft perturbations (e.g., from optimizers) in  continuous domain;
  \item BILVNC, Planet, MIPVerify,  DeepZ, DeepPoly, DeepGo, ReluVal, and DSGMK do not provide any parameters to constrain real adversarial examples
  so that the discretization error cannot be minimized.
\end{itemize}

The remaining 4 methods DeepGauge, SModel, PMG and FD
actually leverage other attack methods
such as (FGSM, BIM, JSMA, and C\&W). Therefore, the impacts of
the discretization problem on their methods rely upon other attacks.

\medskip
\noindent
{\bf Discussion.}
After an in-depth analysis of 35 existing works,
we found that 34 works craft adversarial
example in continuous domains,
29 works are affected by
the discretization problem,
and 23 works do not provide hyper-parameters to avoid the discretization problem.
As aforementioned, real adversarial examples may be damaged
when transform them back into valid images, due to the discretization problem, hence fail to launch attacks.
Besides this, there are other severe consequences:
(1) the black-box methods such as
ZOO, AutoZOOM, GenAttack and SafeCV downgrade to gray-box ones, as
they directly invoke the normalization of the integer classification systems;
(2) the verification methods such as BILVNC,
MIPVerify, Planet and ReluVal that are claimed complete are
only limited to real image classifiers, and become incomplete
on practical image classification systems that are indeed integer image classifiers;
and (3) the verification methods such as BILVNC, MIPVerify, Planet, ReluVal, DeepZ, DeepPoly, DeepGo and DSGMK may craft
spurious adversarial examples and fail to prove robustness of integer image classifiers.

Moreover, during our study,
we found there are lots of Github issues, e.g.,~\cite{issue1,issue2,issue3},
asking why adversarial examples are damaged after saving.
Users might doubt whether implementations are correct or images are saved in a correct way.
According to our findings, it is due to the discretization problem.

\subsection{Empirical Study}\label{sec:Lempstudy}
% We conduct an empirical study on 20 representative open source tools (listed in the first column in Table~\ref{tab:TheExpofDP}),
% in an attempt to understand the impacts of the discretization problem in practice.
% For ease of experiments, we sometimes use the implementations in toolkits (mentioned in the first column in Table~\ref{tab:TheExpofDP}) instead of the original tools,
% and we assume that they are same or better than  their  original implementations.

We conduct an empirical study on 20 representative methods in Table~\ref{tab:TheExpofDP}
whose source code is publicly available,
in an attempt to understand the impacts of the discretization problem in practice.

% in their original tools or the code maintain by the author,
%except for L-BFGS which we use the toolkit Cleverhans~\cite{cleverhans18}
%replaced as the raw paper did not provide open source code link or their code cannot attack in scale.

\begin{landscape}\centering
\begin{table}[t]
 \small
\caption{Experiment results on the discretization problem, where \emph{$\ast$} means the target model in the corresponding tool;
\emph{$\sharp$} means that their tools do not have any target models and we choose widely used target models from Tensorflow or Keras;
% \emph{MSE} means mean-square error,
and \emph{Default} means default input parameters}
\label{tab:TheExpofDP}
\begin{tabular}{|l||c|c|c|c|c|c|c|}

\toprule
{\bf Method}    & {\bf SR}   & {\bf TSR}    & {\bf GAP}   & {\bf Dataset}    & {\bf Model}  &  {\bf Default} &  {\bf Note}                       \\ \midrule% unit of MSE is e-5

FGSM~\cite{GSS15} & 98.61\% & 98.58\%  & 0.03\%  & MNIST  & LeNet-1$^\sharp$  &\cmark  & 10000 images   \\ \midrule  \rowcolor{mygray}

BIM~\cite{KGB17}  & 100\%     & 100\%      & 0\%    & ImageNet   & Inception-v3$^\sharp$ &\cmark & -
\\  \midrule

MBIM~\cite{DLPSZHL18} & 100\%   & 100\%    & 0\%  & ImageNet   & Inception-v3$^\sharp$ &\cmark & -
\\ \midrule   \rowcolor{mygray}

JSMA~\cite{PMJFCS16} & 96\%  & 96\%     & 0\%   & ImageNet      & VGG19$^\sharp$  &\cmark & -
\\ \midrule

L-BFGS~\cite{PE16}  & 100\%      & 77\%      & 23\%     & ImageNet    & Inception-v3$^\sharp$ &\cmark &  -
\\    \midrule  \rowcolor{mygray}
% Foolbox

% C\&W-$\textbf{L}_2$~\cite{CW17b} & 100\% &10\% & 90\%  &ImageNet   & Inception-v3 &  $\kappa$ = 0    \\ \hline
C\&W-$\textbf{L}_2$~\cite{CW17b} & 100\% &10\% & \textbf{90\%}  & ImageNet   & Inception-v3$^\ast$ &\cmark & -    \\ \midrule

DeepFool~\cite{MFF16}  & 100\%             & 23\%           & \textbf{77\%}      & ImageNet      & ResNet34$^\ast$  &\cmark & -      \\ \midrule  \rowcolor{mygray}

DeepXplore~\cite{PCYJ17}& 65\%   & 28\%   & \textbf{56.92\%}  & ImageNet    & ResNet50, VGG16\&19$^\ast$ &\cmark & Generate examples with 100 seeds  \\ \midrule

DeepConcolic~\cite{SWRHKK18}    & 2\%      & 2\%          & 0\%  & MNIST      &  mnist\_complicated.h5$^\ast$  &\cmark &  10000 images with criterion=`nc' \\ \midrule \rowcolor{mygray}

% One-Pixel~\cite{SU17} & 3\%             & 3\%          & 0\%        & MNIST    & LeNet-1    &  max\_pixels=9 with 100 images      \\ \hline

ZOO~\cite{CZSYH17}& 58\%            & 6\%          & \textbf{89.66\%}   & ImageNet    & Inception-v3$^\ast$ &\cmark & -  \\ \midrule   \rowcolor{mygray}

DBA~\cite{BRB18} & 100\%         & 28\%          & \textbf{72\%}     & ImageNet       & VGG19$^\sharp$ &\cmark & -   \\ \midrule

NES-PGD~\cite{IEAL18} & 100\%   & 53\%   & 47\% & ImageNet    & Inception-v3$^\ast$ &\cmark & - \\  \midrule   \rowcolor{mygray}

Bandits~\cite{IEM2018PriorCB}& 94\%   & 11\%   & \textbf{88.3\%}   & ImageNet  & Inception-v3$^\ast$ &\cmark & -  \\ \midrule	

GenAttack~\cite{ASCZHS19} & 100\%   & 91\%   & 9\%   & ImageNet  & Inception-v3$^\ast$ &\cmark & -  \\ \midrule

DLV~\cite{HKWW17}          & 90\%         & 90\%          & 0\%      & MNIST      & NoName$^\ast$ &\cmark &  20 images  \\ \midrule   \rowcolor{mygray}
% SR not 100\% as time out, 20 images  \\ \hline   \rowcolor{mygray}

Planet~\cite{Ehl17}         & 100\%        & 46\%          & \textbf{54\%}     & MNIST      & testNetworkB.rlv$^\ast$ &\cmark & Use `GIVE' model obtain 20 images  \\ \midrule

MIPVerify~\cite{TXT18}            & 42\%       & 0\%       & \textbf{100\%}  & MNIST & MNIST.n1$^\ast$ &\cmark & Quickstart demo with 100 images         \\ \hline    \rowcolor{mygray}

DeepPoly~\cite{SGPV19}       & 45\%      & 44\%      & 2.22\% &  MNIST & convBigRELU\_DiffAI$^\ast$ &\cmark & Gap between $\epsilon=0.3$ and $\epsilon=76/255$  \\ \midrule  % with 100 images

DeepGo~\cite{RHK18}         & 25.4\% & 25.2\% & 0.78\%    & MNIST     & NoName$^\ast$  &\cmark & Crafted 1000 images from 1 image \\ \midrule   \rowcolor{mygray} % with featureIndexZ = 2

SafeCV~\cite{WHK18}         & 100\%      & 100\%     & 0\% & MNIST & NoName$^\ast$ &\cmark &  100 images        \\ \bottomrule

%& 100\%         & 91\%          & 9\% (1.61e-03)        & ImageNet       & Inception-v3 &  $\epsilon = 0.05, \alpha \approx 0.15$    				 \\ \rowcolor{mygray}
%%\multirow{-2}{*}{GenAttack~\cite{ASCZHS19}}\cellcolor{white} &  99\%         &46  \%          &  53.54 \%  (4.39e-04)       & ImageNet       & Inception-v3 &  $\epsilon = 10/255, \alpha = 0.1$    		\\ \hline

%C\&W-$\textbf{L}_2$+GS~\cite{CW17b}    & 100\% &100\%   & 0\%     &MNIST   & Original & \tabincell{l}{bin\_search=3, init\_const=1,  greedy-enabled, 900 images} \\ \hline
%C\&W-$\textbf{L}_2$~\cite{CW17b}\cellcolor{white} & 100\% &22.89\% & 77.11\% & 4.57e-3  &MNIST   & Original &  bin\_search=9, init\_const=1e-3, greedy-disabled, 900 images  \\ \rowcolor{mygray}
%greedy\cellcolor{white} & 100\% &99.67\% & 0.22\% & 4.6e-3  &MNIST   & Original & bin\_search=9, init\_const=1e-3, greedy-enabled, 900 images\\ \rowcolor{mygray}
%search\cellcolor{white}  & 100\% &100\%   & 0\% & 4.56e-3      &MNIST   & Original & \tabincell{l}{bin\_search=3, init\_const=1,  greedy-enabled, 900 images} \\ \hline
\end{tabular}
\end{table}
\end{landscape}

We consider the following two research questions:
\begin{description}
%  \item[RQ1:] Does perturbation step size matter w.r.t. the discretization problem?
  \item[{\bf RQ1}:] To what extent does the discretization problem affect the attack success rate?
  \item[{\bf RQ2}:] Can the discretization problem be avoided or alleviated by tuning input parameters?
\end{description}

%\subsubsection{Dataset, Metrics and Setting}

\medskip
\noindent{\bf Setting.}
In our experiments, we use the official implementations of the authors.
Due to the diversity of these tools,
the dataset and setting may be different. %, details are listed in Table~\ref{tab:TheExpofDP}).
We manage to be consistent with the original environments in their raw papers,
attack the target models provided by the tools,
and conduct targeted attacks unless the tools are designated for untargeted attacks.
For verification tools that cannot directly attack the model,
we evaluate them by analyzing the generated counterexamples.
Although, we do not change their settings deliberately to get exaggerative results,
we should emphasize that the comparison between these tools may be unfair,
\emph{our main goal is to understand their own tools.}

\medskip
\noindent{\bf Dataset.}
We use two popular image datasets:
MNIST~\cite{MNIST98}
and ImageNet~\cite{DDSLL009}.
%The size of images in ImageNet is around $299\times299\times3$ or $224\times224\times3$ pixels.
ImageNet contains over $10000000$ images with $1000$ classes.
We randomly choose 100 classes from which
we randomly choose 1 image per class that can be correctly classified by four classifiers in Keras: ResNet50, Inception-v3, VGG16 and VGG19.
For MNIST images, the numbers of used images are shown in the last column in Table~\ref{tab:TheExpofDP},
which depends on the efficiency of the tool under test.

\medskip
\noindent{\bf Metrics.}
We introduce three metrics to evaluate the impacts of the discretization problem.
Let $N$ denote the number of input images under test,
$N_{v}$ denote the number of successfully crafted real adversarial examples,
and $N_{i}$ denote the number of integer adversarial examples after the denormalization post-processing,
\begin{itemize}%[noitemsep,topsep=0pt,leftmargin=*]
  \item {Success Rate (SR)}
   is calculated as  $\frac{N_{v}}{N}$,
  \item {True Success Rate (TSR)}
is calculated as  $\frac{N_{i}}{N}$,
  \item {GAP between SR and TSR } is calculated as $\frac{SR-TSR}{SR}$.
\end{itemize}
To compute $N_i$, we use the denormalizer provided by the corresponding tools.

\medskip
\subsubsection{\bf RQ1}
To answer this research question, we conduct experiments
using default input parameters in their raw papers or tools, which have been fine-turned for effectiveness by corresponding authors and widely used by existing works.
The results are shown in Table~\ref{tab:TheExpofDP}.

We can observe that 14 out of 20 tools are affected by the discretization problem.
Their gaps range from 0.03\% to 100\%.
In more detail,
8 tools have gaps exceeding 50\%
including white-box testing tools (C\&W-$\textbf{L}_2$, DeepFool, DeepXplore),
black-box testing tools (ZOO, DBA and Bandits) and verification tools (Planet and MIPVerify).
Among them, 6 tools have gaps exceeding 70\%.
This demonstrates that if attackers do not pay attention on the discretization problem,
they will be likely to generate real adversarial examples which will
be damaged after transforming them back into the discrete domain.

There are only 6 out of 20 tools that do not have any gaps including BIM, MBIM, JSMA, DeepConcolic, DLV and SafeCV.
These results are largely consistent with our theoretical study.
%We remark that DeepConcolic only have 2\% attack success rate.

\medskip
\begin{tcolorbox}[left=0mm,right=0mm,top=0mm,bottom=0mm]
\textbf{Answering RQ1:} The results on 20 tools show that most of them are affected by the discretization problem.
There are 8 tools whose gap exceeds 50\%, and 6 tools
whose gap exceeds 70\%, and only 6 tools do not have any gaps.
\end{tcolorbox}

%As aforementioned, FGSM crafts adversarial examples using a relative larger step size $0.3$ in one step,
%therefore, the adversarial examples crafted by FGSM are more robust.

\subsubsection{\bf RQ2}\label{sec:tuning-para}
To answer this research question, we propose different strategies to tune input parameters for these $14$ tools whose gap is not $0$ in RQ1.
According to our findings in
theoretical study, we distinguish these tools by
whether the discretization problem can be easily and directly
avoided by tuning input parameters.
Remark that we do not investigate how to 
modify their implementations and methods 
by taking the discretization problem into account.
First, it is a tedious and error-prone process.
Second, modifying their implementations may greatly under-estimate their effectiveness
and efficiency, as pointed out by Carlini~\cite{Car19}, hence less convincing.

\medskip
\noindent{\bf In theoretical study,
the discretization problem can be easily and directly avoided by tuning input parameters.}
Based on the results in Table~\ref{tab:TheExpofDP},
we can observe that only FGSM has non-zero gap and
its discretization problem can be easily and directly
avoided by tuning input parameters.
The default perturbation step size
$\epsilon$ used in Table~\ref{tab:TheExpofDP} is $0.3$.
Therefore, we revise $\epsilon$ to $76/255$ in order to avoid the discretization problem.
%The result is shown in Table~\ref{tab:TheExpofDPtuning}.
Then, the gap is decreased to $0$ with TSR $98.55\%$. This confirms our theoretical findings.

To illustrate the importance of controllable perturbation step sizes,
we also test the implementations of BIM and MBIM in other toolkits, such as Foolbox~\cite{RBB17}.
Different from the raw implementation of these tools,
Foolbox provides a binary search by default.
The binary search is performed between the original clean input and the crafted adversarial image, intending
to find adversarial boundary.
It has been adopted in recent attacks, e.g., ~\cite{BRKUB19,CurlsWhey2019CVPR}.
However, if the binary search is implemented without taking into the discretization problem account such as BIM and MBIM in Foolbox,
the perturbation step size will become uncontrollable.
We use the same input parameters of BIM and MBIM as in RQ1, exception that
the binary search is enabled (default in Foolbox). %The results in shown in Table~\ref{tab:TheExpofDPtuning}.
Compared to the results in Table~\ref{tab:TheExpofDP},
the gaps of both BIM and MBIM increase from 0\% to 90\%.
This shows that attackers should pay more attention on input parameters
even the discretization problem is avoidable.

\medskip
\noindent{\bf  In theoretical study, the discretization problem cannot be easily and directly
avoided by tuning input parameters.}
Based on the results in Table~\ref{tab:TheExpofDP},
there remain 13 tools whose gaps are non-zero,
and the discretization problem cannot be easily and directly
avoided by tuning input parameters.
We do our best to fine-turn input parameters of
those tools aimed at increasing TSR and decreasing gap.

First of all, as discussed in theoretical study, the verification tools (i.e., Planet, MIPVerify, DeepPoly and DeepGo)
do not provide any parameters to constrain real adversarial examples so that the discretization error could be minimized, we cannot tune input parameters
of those tools.
For the other 9 test-based tools (i.e., white-box attacks L-BFGS, C\&W, DeepFool and DeepXplore,
and black-box attacks ZOO, DBA, NES-PGD, Bandits and GenAttack),
we adopt the following three strategies to alleviate
the discretization problem:
\begin{itemize}
  \item[S1:] forbidding adaptive perturbation step size: aims at controlling perturbation step sizes. NES-PGD, DBA, Bandits and GenAttack provide such adaptive mechanism.
  \item[S2:] increasing overall perturbations: aims at minimizing the ratio of discretization error against the overall perturbations. L-BFGS, DeepFool, DeepXplore and DBA
provide input parameters related to this strategy.
  \item[S3:] enhancing strength/confidence of adversarial examples: aims at enhancing the robustness of real adversarial sample.
  C\&W and ZOO provide input parameter related to confidence.
\end{itemize}

% \begin{table}[t]
% \caption{Results of NES-PGD and GenAttack}
% \label{tab:restunes1}\setlength{\tabcolsep}{2pt}	
% \centering
% %\vspace{-5pt}
% \scalebox{1}{
% \begin{tabular}{cccccc}
% \multicolumn{5}{l}{NES-PGD~\cite{IEAL18} attack on ImageNet+Inception-v3} \\
% \multicolumn{5}{l}{Version1 (default): $\epsilon = 0.05, \alpha \in [1e-2, 5e-5]$} \\
% \multicolumn{2}{l}{Version2: $\epsilon = 10/255 $} &
% \multicolumn{3}{l}{Version3: $\epsilon = 10/255, \alpha = 1/255$} \\
% \hline
% Version  & SR & TSR & $L_\infty$-distance & Mean Queries& Discr. Error \\
% \hline
% 1  & 100\% & 53\%  & 0.05$\times$255 & 12470 &$\approx 0.5$\\
% 2  & 100\% & 47\%  & 10  & 16168    & $\approx 0.5$   \\
% 3  & 74\%  & 73\%  & 10  & 214552    & $\approx 0.16$  \\
% \hline
% \\
% \multicolumn{5}{l}{GenAttack~\cite{ASCZHS19} attack on ImageNet+Inception-v3} \\
% \multicolumn{5}{l}{Version1(default): $\epsilon = 0.05, \alpha \approx 0.15$, adaptive=True} \\
% \multicolumn{5}{l}{Version2: $\epsilon = 10/255, \alpha \approx 0.1$, adaptive=True} \\
% \multicolumn{5}{l}{Version3: $\epsilon = 10/255, \alpha = 0.1$, adaptive=False} \\
% \hline
% Version  & SR & TSR & $L_\infty$-distance  & Mean Queries & Discr. Error \\
% \hline
% 1  & 100\% & 91\% & 0.05$\times$255 &  24728& $\approx 0.5$\\
% 2  & 97\%  & 56\% & 10 &  33273 &$\approx 0.5$\\
% 3  & 99\%  & 46\% & 10 &   45576 & $\approx 0.5$\\
% \hline
% \end{tabular}}%\vspace{-2mm}
% \end{table}

After tuning input parameters,
none of them is able to  eliminate the discretization errors absolutely.
%namely, the maximum rounding error.
%when transforming a real adversarial image back into the discrete domain is non-zero. These results also consist with our theoretical study.

In terms of TSR, we found that:
\begin{itemize}
\item By applying  S1, the TSR of NES-PGD and DBA can increase, % at the cost of larger perturbations or higher query times,
but the TSR of Bandits and GenAttack cannot;
\item By applying  S2, the TSR of DBA can increase, %at the cost of larger perturbations,
but the TSR of DeepXplore, L-BFGS and DeepFool cannot;
\item By applying  S3, the TSR of C\&W-$\textbf{L}_2$ can increase,
but the TSR of ZOO cannot.
\end{itemize}
This demonstrates that our strategies are able to increase TSR for 3 tools, but fail to increase TSR for the other 6 tools.
However, these strategies also bring some side effects,
namely, increasing either overall perturbations in terms of Mean Square Error (MSE)
or the number of query times, hence sacrificing attack efficiency and imperceptibility of adversarial samples.
Due to limited space, detailed statistic is given in Appendix~\ref{sec:app1}.

\medskip
\begin{tcolorbox}[left=0mm,right=0mm,top=0mm,bottom=0mm]
\textbf{Answering RQ2:}
According to our experiences,
among 14 tools that are affected by the
discretization problem,
only 1 tool, FGSM, can definitely avoid the
discretization problem by tuning input parameters,
% 10 tools can neither avoid nor alleviate the
% discretization problem by tuning input parameters,
and only 3 tools can alleviate the
discretization problem by tuning input parameters
at the cost of attack efficiency or imperceptibility.
\end{tcolorbox}

%All the tools that did not consider the discretization problem have, more and less, gaps.
%These tools implemented either gradient-based (such as
%BIM, MBIM, C\&W, and DeepXplore) or non-gradient-based
%(such as L-BFGS, DeepFool and ZOO), or verification-based (such as MIPVerify) methods.
%Also, some of them are white-box attacks (such as
%BIM, MBIM, C\&W, DeepXplore and MIPVerify)
%and some of them are black-box attacks (such as ZOO, NES-PGD and DBA).
%Therefore, the discretization problem is ubiquitous no matter test based or verification based,
%and black-box or white-box.

\medskip

\noindent{\bf Discussion.}
Our empirical study reveals that the discretization problem is more severe than originally thought
in practice, in conformance with the results of our theoretical study.
According to our experimental results, the attack results in published works
may not be as good as those reported in raw papers.
For instance, DeepFool assumed that
the classifier ${f}_{t}$ in continuous domain
is the same as the classifier in the concrete domain ${g}_{t}$
which contradicts to our empirical result, e.g. it has gap $77\%$ in Table~\ref{tab:TheExpofDP}.
We believe it is important to highlight the potential impacts of the discretization problem,
and by no means invalidate existing methods or their importance and contributions.

It is worth to note that Carlini and Wagner~\cite{CW17b} proposed a greedy search based algorithm
to alleviate the discretization problem.
We conduct an experiment on the greedy search based version of C\&W-$\textbf{L}_2$ which are obtained from Carlini.
In our experiment, we use input parameters
recommended by Carlini for MNIST images.
We found that the greedy search based algorithm significantly
improves TSR and reduces gaps without increasing
distortions of crafted adversarial examples.
This demonstrates that the greedy search based algorithm is a
solution to alleviate the discretization problem when one cannot precisely control perturbation step sizes
by adjusting input parameters.
However, due to the fact that the greedy
search based algorithm leverages gradients of targeted networks frequently,
it is difficult to integrate it into black-box attacks.

According to our findings, we suggest that:
(1) attack success rate should be measured
using integer adversarial examples instead of real adversarial examples;
(2)  it is vital to pay more attention
to perturbation step sizes that can be controlled
by input parameters;
and (3) it is better to revise the implementations of
the tools that cannot easily avoid the discretization problem by tuning input parameters
if one wants to achieve higher TSR but do not sacrifice
the attack efficiency and imperceptibility of adversarial samples.

%% file: methodology.tex
\section{An Approach for Black-box Attack}
\label{sec:method}
According to our study in Section~\ref{sec:DPprobleminWild},
there lacks an effective and efficient \emph{integer} adversarial example attack
in black-box scenario.
%
%To do this, % integer adversarial examples in black-box scenario,
%one can only query the classifier % for input images
%and get
%the probabilities of top-k classes for each input.
%Then, we have to compute integer perturbations of images in the discrete domain,
%which always results in integer adversarial examples.
As a first step towards addressing this problem,
%in this section,
we propose a novel black-box algorithm for %crafting integer adversarial examples for
both targeted and untargeted attacks by presenting a \emph{classification model-based} \textsc{d}erivative-\textsc{f}ree discrete \textsc{o}ptimization (DFO) method. %~\cite{yu2016derivative}.
This type of DFO methods has been widely used to solve complex optimization tasks in a sampling-feedback-style.
It does not rely on the gradient of the objective function, but instead,
learns from samples of the search space.
Therefore, it is suitable for optimizing functions that are non-differentiable,
%with many local minima,
or even unknown but only testable.
Furthermore, it was shown by Yu et al.~\cite{yu2016derivative}
that it is not only superior
to many state-of-the-art DFO methods (e.g., genetic algorithm, Bayesian optimization and cross-entropy
method), but also stable.
We refer readers to~\cite{yu2016derivative} for the advantages of classification model-based DFO methods.

In the rest of this section, we first introduce our approach framework, then present the formulation
and our algorithm.

\medskip
\noindent
{\bf Threat model}.
In our black-box scenario,
we assume that the adversary does not have any access to any details (e.g.,  normalization,
architecture, parameters and training data) of the target classifier,
but he/she knows the input format
of the target classifier and has access to
the probabilities (or confidences) of all classes for each input image which is
a widely used assumption even in black-box scenario~\cite{SL14,xu2016automatically,PMGJCS17,ilyas2017query,BHLS18}.
The distortion of adversarial examples is measured by the
$\mathds{L}_\infty$ distance metric.

\subsection{Framework of \tool}

%We formalize the computation of adversarial image examples as a black-box discrete optimization problem
%constrained with a distance threshold of $\mathds{L}_\infty$ distance metric.
%However, this discrete optimization problem cannot be solved using gradient-based methods, as
%the model is non-continuous.
%To overcome this challenge, we leverage a derivative-free discrete optimization method that
%does not rely on the gradient of the objective function, but instead,
%learns from samples of the search space.
%It is suitable for optimizing functions that are non-differentiable,
%with many local minima, or even unknown but only testable.

\begin{figure}[t]
  \centering
  \includegraphics[width=0.85\textwidth]{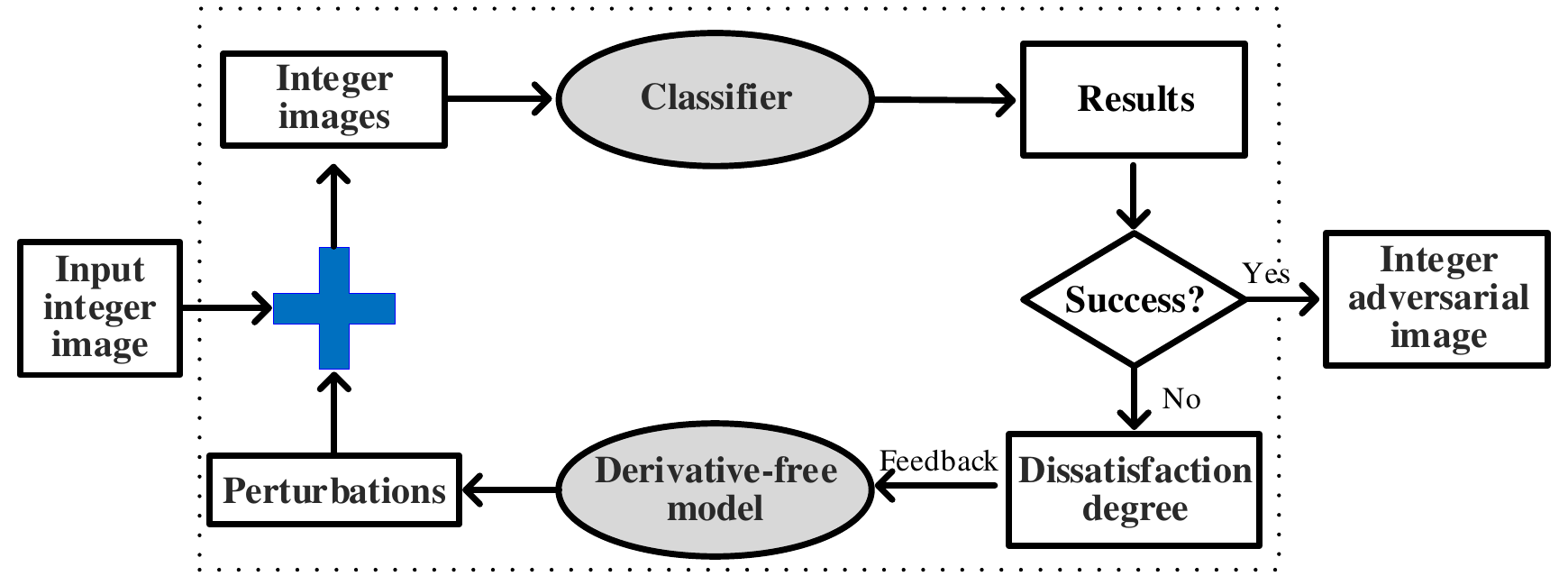}
    \caption{Framework of our approach \tool.}
  \label{fig:frame}
\end{figure}

Figure~\ref{fig:frame} shows the framework of
our approach named \tool, standing for \textbf{D}erivative-\textbf{F}ree \textbf{A}ttack.
Given an integer image, \tool directly searches an  adversarial image in a (discrete integer) search space specified by the maximum $\mathds{L}_\infty$ distance.

In principle, DFA first samples some perturbations from the search space
and then repeats the following procedure until an integer adversarial example is found.
%
%adds them onto the input image, producing candidate integer images. % by adding these perturbations onto the original image.
%Then, %starting from the set of crafted images,
%it repeats the following procedure until an integer adversarial example is found. %(i.e., an image with minimal dissatisfaction degree that measures how far is a specific perturbation from a success attack) or the number of iterations is reached.
During each iteration, \tool queries the target classifier to measure
the images (perturbations added onto the input image) via a given \emph{dissatisfaction degree function} which predicates how far is an image from a success attack.
The perturbations is partitioned into two parts w.r.t. \emph{dissatisfaction degrees}:
perturbations yielding \emph{high} dissatisfaction degrees
and perturbations yielding \emph{low} dissatisfaction degrees.
The search space is refined into a \emph{small sub-space} according to the partitions of perturbations.
New perturbations are sampled from the refined sub-space. Together with old perturbations,
a set of best-so-far perturbations is selected according to their dissatisfaction degrees.
Finally, the procedure is repeated on the best-so-far perturbations
which will be used to refine the sub-space again.

\subsection{Formulation}
\label{sec:Formulation}

%After the previous description of the framework, our attack algorithm is as follows: having chosen the derivative-free model with a dissatisfaction-degree function, we create an altered image by generating a image perturbation to add to the original image, which is the input of the neural network. Repeat this process with updating the model by the dissatisfaction-degree function until that an adversary example has been found.

We formalize the integer adversarial example searching problem
as a derivative-free discrete optimization problem by defining the dissatisfaction degree functions.
We first introduce some notations.

Let us fix a classifier $f_t:\mathds{D}\rightarrow \mathds{C}_t$ for some
image classification task $t$ and an integer number $\epsilon$ denoting the maximum $\mathds{L}_\infty$ distance.
We denote by $\Prr(\vec{d})$ the vector of probabilities on
the image $\vec{d}$ and
by $\Prr(\vec{d},c)$ the probability that the image $\vec{d}$ is classified to the class $c \in\mathds{C}_t$.
For a given integer $j$ such taht $1\leq j\leq |\mathds{C}_t|$, we denote
by $\Ttop_j(\vec{d})$ the $j$-th largest probability in $\Prr({\vec{d}})$
and $\Ttop_j^{\ell}(\vec{d})$ the class whose probability is $\Ttop_j(\vec{d})$ .
Obviously, $\Ttop_1^{\ell}(\vec{d})=f_t(\vec{d})$.

We define the initial search space $\Delta$ of perturbations
as a discrete integer domain $\mathds{N}_{[-\epsilon,\epsilon]}^{w\times h \times {ch}}$.
Specifically, the discrete domain $\Delta$ is a two-dimensional array such that for each
coordinate $p\in P=w\times h \times {ch}$,
$\Delta[p][{\tt low}]$ and $\Delta[p][{\tt high}]$ (such that $\Delta[p][{\tt high}]\geq \Delta[p][{\tt low}]$)
are integer numbers respectively denoting the lower and upper bound of
the value at the coordinate $p$.
Therefore, $\Delta$ denotes a set of perturbations
such that $\delta\in\Delta$ if and only if $\Delta[p][{\tt low}]\leq \delta[p]\leq \Delta[p][{\tt high}]$ for all coordinates $p\in P$.
The search space $\Delta$ will be refined into small sub-spaces by increasing
lower bound $\Delta[p][{\tt low}]$ or decreasing upper bound  $\Delta[p][{\tt high}]$ for choosing coordinates $p$ in our algorithm.

%$\Delta[p][q]$ indicates the lower bound/upper bound of the space for certain coordinate $p$.
%Given a perturbation $\delta\in\Delta$, and a search space $\Delta$, we say $\delta\models \Delta$, if for every coordinate $p\in P$, $\delta[p]\in$ $\Delta[\vec{d}[p], 0] ,\Delta[\vec{d}[p], 1] $

 % initially. %$region$ defines the limitation of the value of the perturbation.

Given a perturbation $\delta\in\Delta$, we denote by $\vec{d}\oplus\delta$,
the valid image after adding the perturbation $\delta$ onto the image $\vec{d}$,
namely, for every coordinate $p\in P$:
\[(\vec{d}\oplus\delta)[p]:=\left\{
                            \begin{array}{ll}
                              \vec{d}[p]+\delta[p], & \hbox{if } 0\leq \vec{d}[p]+\delta[p]\leq 255; \\
                              0, & \hbox{if } \vec{d}[p]+\delta[p]<0; \\
                              255, & \hbox{if } \vec{d}[p]+\delta[p]>255.
                            \end{array}
                          \right.\]

The \emph{integer adversarial example searching problem} with respect to the maximum $\mathds{L}_\infty$ distance $\epsilon$
is to find some perturbation $\delta\in\Delta$
such that:
\begin{itemize}%[leftmargin=*]
  \item $\Ttop_1^{\ell}(\vec{d}\oplus\delta)\neq f_t(\vec{d})$
for untargeted attack;
  \item $\Ttop_1^{\ell}(\vec{d}\oplus\delta)=c$
for targeted attack with a target class $c$.
\end{itemize}

We solve the integer adversarial example searching problem by reduction to
a derivative-free discrete optimization problem. The reduction
is given by defining an optimization goal which is characterized
by \emph{dissatisfaction-degree functions}.
We first consider the untargeted case.

The goal of untargeted attack is to find some perturbation $\delta\in\Delta$
such that $\Ttop_1^{\ell}(\vec{d}\oplus\delta)\neq f_t(\vec{d})$.
To do this, we %minimize the probability of
%$\vec{d}\oplus\delta$ being classified as the class $f_t(\vec{d})$
%and
%minimize the distance between the probability of the image $\vec{d}\oplus\delta$ being classified as the class $f_t(\vec{d})$
maximize the current probability of the image $\vec{d}\oplus\delta$ being classified as
the class $\Ttop_2^{\ell}(\vec{d}\oplus\delta)$ (i.e., the current class with second largest probability, which may change w.r.t. different $\delta$) until
the image is able to successfully mislead the classifier.
Therefore, we define
the \emph{dissatisfaction-degree function} for untargeted attack, denoted by $D_{\tt ua}(\cdot,\cdot)$, as follows:  %\footnote{In order to accelerate the searching of adversarial examples, we use the  logarithmic values of these probabilities in the computation.}:

\begin{itemize}%[leftmargin=*]
  \item $D_{\tt ua}(\vec{d},\delta):=0$, if $\Ttop_1^{\ell}(\vec{d}\oplus\delta)\neq f_t(\vec{d})$;
  \item $D_{\tt ua}(\vec{d},\delta):=1-\Ttop_2(\vec{d}\oplus\delta)$, otherwise.
\end{itemize}

In this function, if the attack has succeeded, the perturbation $\delta$ is ``satisfying'', then the value of the dissatisfaction-degree becomes $0$.
Otherwise, we return the distance between $1$ and the currently reported second largest probability, which is in the range of $[0,1]$,
indicating how far it is from 1. Clearly, in this case, the distance is definitely positive.
To this end, our goal is to find a perturbation $\delta$ such that the dissatisfaction-degree is $0$. %We need to note that, during the search, the class with second largest probability may change

For targeted attack with the target class $c$,
instead of maximizing the probability of $\vec{d}\oplus\delta$ being classified as
the class $\Ttop_2^{\ell}(\vec{d}\oplus\delta)$, we maximize the probability of the image $\vec{d}\oplus\delta$ being classified as
$c$. Hence,
the \emph{dissatisfaction-degree function},
denoted by $D_{\tt ta}(\cdot,\cdot)$, is defined as follows:

\begin{itemize}%[leftmargin=*]
  \item $D_{\tt ta}(\vec{d},\delta):=0$, if $\Ttop_1^{\ell}(\vec{d}\oplus\delta)=c$;
  \item $D_{\tt ta}(\vec{d},\delta):=1-\Prr((\vec{d}\oplus\delta),c)$, otherwise.
%   \item $D_{\tt ta}(\vec{d},\delta):=\log(\Ttop_1(\vec{d}\oplus\delta))-\log(\Prr(\vec{d}\oplus\delta),c)$, otherwise.
\end{itemize}

Now, the integer adversarial example searching problem
is reduced to the minimization problem of the dissatisfaction-degree functions.

\begin{algorithm}[t]
	\renewcommand{\algorithmicrequire}{\textbf{Input:}}
	\renewcommand{\algorithmicensure}{\textbf{Output:}}
\small
	\caption{A DFO-based algorithm}
	\label{alg:1}
	\begin{algorithmic}[1]
		\REQUIRE classifier under attack $f_t:\mathds{D}\rightarrow \mathds{C}_t$, integer image $\vec{d}\in \mathds{D}$, number of iterations $T\in\mathds{N}$, \\
         ranking threshold $k\in\mathds{N}$, sample size $s\in\mathds{N}$, maximum $\mathds{L}_\infty$ distance $\epsilon\in\mathds{N}$,\\
        the number of coordinates to be changed in each refinement process $u\in\mathds{N}$,\\
        	dissatisfaction-degree (d.d.) function $D$\\
        \ENSURE optimized perturbation $\tilde{x}$
       % \STATE \emph{/*main procedure*/}
         \STATE $\Delta= \mathds{N}_{[-\epsilon,\epsilon]}^{w\times h \times {ch}}$;
         \STATE $B_0= \{\delta_1,...,\delta_{s+k}\}$  sampled from $\Delta$;
           \emph{   // initial collection}
         %\STATE Compute images $(\vec{d}\oplus\delta_i)$ for $1\leq i\leq s+k$;
        \STATE Evaluate the dissatisfaction-degree $D(\vec{d},\delta_i)$  for $1\leq i\leq s+k$;
        \STATE $\tilde{x} = {\tt argmin}_{\delta\in B_0}D(\vec{d},\delta)$; \emph{// select the best-so-far sample}
        \FOR{$t=1$ \TO $T$}
            \IF {$D(\vec{d},\tilde{x}) = 0$} \STATE  {\bf break}; \emph{ // find an adversarial example}
            \ENDIF
            \STATE $B_{t-1}^{+} =$ smallest-k solutions in $B_{t-1}$ in terms of d.d.;
            \STATE $B_{t-1}^{-} = B_{t-1}-B_{t-1}^{+}$;
            \STATE $B=\emptyset$;
            \FOR{$i=1$ \TO $s$}
            \STATE \emph{// Refine the  space $\Delta$ into a small one by $B_{t-1}^{+}$ and $ B_{t-1}^{-}$}
                \STATE Randomly select a sample $b^{+}$ from the positive  set $B_{t-1}^{+}$;
                \STATE $Y=\emptyset$;
                \FOR{$j=1$ \TO $u$}
                    \STATE Randomly select a coordinate $p$ from $P=w\times h\times ch$;
                    \STATE $Y = Y\cup \{p\}$;
                    \STATE ${\tt ge}:=\{ b \in B_{t-1}^{-}\mid b[p]>b^{+}[p]\}$;
                    \STATE ${\tt le}:=\{ b \in B_{t-1}^{-}\mid b[p]<b^{+}[p]\}$;
                    \IF {$|{\tt ge}| > |{\tt le}|$}
                        \STATE ${\tt minVal} = {\tt min}_{b\in {\tt ge}}b[p]$;
                        \STATE Randomly select an integer $r$ from ${\tt minVal}$ to  $b^{+}[p]$;
                        \STATE $\Delta[p][{\tt high}]=r$;  \emph{// decrease the upper bound at $p$}
                    \ELSE
                        \STATE ${\tt maxVal} = {\tt max}_{b\in {\tt le}}b[p]$;
                        \STATE Randomly select an integer $r$ from $b^{+}[p]$ to ${\tt maxVal}$;
                        \STATE $\Delta[p][{\tt low}]=r$;
                        \emph{// increase the lower bound at $p$}
                    \ENDIF
                \ENDFOR
                \STATE $b' ={\tt Copy \ of \ } b^{+}$;
                \FOR{$p \in Y$} \STATE \emph{// Sample in the refined the search space $\Delta$}
                    \STATE Randomly select an integer $r$ from $\Delta[p][{\tt low}]$ to $\Delta[p][{\tt high}]$;
                    \STATE $b'[p] = r$;
                \ENDFOR
                \STATE $B = B\cup \{b'\}$;
                \STATE $\Delta= \mathds{N}_{[-\epsilon,\epsilon]}^{w\times h \times {ch}}$; \emph{// Reset $\Delta$ for next sample to avoid over fitting}
            \ENDFOR
            %\STATE Compute images $(\vec{d}\oplus\delta)$ for all $\delta\in B$;
            \STATE Evaluate the dissatisfaction-degree $D(\vec{d},\delta)$ for all $\delta\in B$;
           % \STATE $B_{t-1}$ rank by $Df$
           \STATE $B_t=$ smallest-$(s+k)$ solutions in $B\cup B_{t-1}$ in terms of dissatisfaction-degree  \emph{// keep the size as $s+k$};
           % \STATE $B_t = the in B_{t-1}\}$     \emph{//keep the size of set $B$}
            \STATE $\tilde{x} = {\tt argmin}_{\delta\in B_t}D(\vec{d},\delta)$;
        \ENDFOR
        \RETURN $\tilde{x}$;
	\end{algorithmic}
\end{algorithm}

\subsection{Algorithm}

Instead of using heuristic search methods, e.g. genetic programming, particle swarm optimization, simulated annealing, to solve the minimization problem of the dissatisfaction-degree functions, %in this section,
we propose a classification model-based  DFO method (shown in Algorithm~\ref{alg:1}).
Different from heuristic search based methods,
our method maintains a classification model during the search to distinguish ``good'' samples from ``bad'' samples. Then,
the search space $\Delta$ will be refined by learning from the samples to help to converge to the best solution.

%With the dissatisfaction-degree function, we can evaluate the quality of the sampled solutions and change the adversary example generation problem into an optimization problem. Thus, we can use DFO method to handle this problem. Our algorithm is shown in Alg.\ref{alg:1}.

In detail, Algorithm~\ref{alg:1} first initializes the search space $\Delta$
according to the given maximum $\mathds{L}_\infty$ distance $\epsilon$ (Line 1).
Then, it randomly selects $(s+k)$ perturbations  (stored as the set $B_0$) from the search space $\Delta$ (Line 2),
where $s$ denotes the sample size during each iteration and $k$ denotes
the ranking threshold.
Next, it computes $(s+k)$ valid images
by adding the perturbations onto the source integer image
$\vec{d}$ and evaluates the dissatisfaction-degree (d.d.)
of the these images using the dissatisfaction-degree function $D$ (Line 3).
The perturbation $\tilde{x}$ with the smallest dissatisfaction-degree
is selected from the set $B_0$ (Line~4).
After that, Algorithm~\ref{alg:1} repeats the following procedure.

For each iteration $t\geq 1$, if the perturbation $\tilde{x}$ suffices to craft an integer adversarial example,
return $\tilde{x}$ (Lines 6-7).
Otherwise, the set $B_{t-1}$ of perturbations is partitioned into two sets:
``positive'' set $B_{t-1}^{+}$ and  ``negative'' set  $B_{t-1}^{-}$,
where $B_{t-1}^{+}$ consists of the
smallest-$k$ perturbations in terms of the dissatisfaction-degree (Lines 8-9).

Based on  $B_{t-1}^{+}$ and  $B_{t-1}^{-}$, Algorithm~\ref{alg:1} refines the search space $\Delta$ into a small sub-space
(Lines 11-34) as follows.
It first randomly selects a sample $b^{+}$ from the positive set $B_{t-1}^{+}$  (Line 13)
and randomly selects $u$ coordinates to refine (Lines 15-27).
For each selected coordinate $p$,
it compares the number of perturbations in $B^{-}_{t-1}$
whose value is larger than
the value of $b^{+}$ at the coordinate $p$ against the number of perturbations in $B^{-}_{t-1}$
whose value is smaller than
the value of $b^{+}$ at the coordinate  $p$.
If the majority of perturbations in $B^{-}_{t-1}$ are larger than $b^{+}$ at the coordinate $p$,
we decrease the upper bound of the search space $\Delta$ (Lines 20-23),
otherwise increase the lower bound (Lines 25-27), at the coordinate $p$.
Once $u$ coordinates have been processed,
we craft a new image $b'$ from $b^{+}$
by reassigning the value of each coordinate $p\in Y$
with the random integer $r$ from $\Delta[p][{\tt low}]$ to $\Delta[p][{\tt high}]$ (Lines 29-33).
The new perturbation $b'$ is added into the set $B$.
Then, the search space $\Delta$ is reset to the original size. We remark that the refining procedure for the next sample will be conducted on the original search space to avoid over fitting.
%:
%a perturbation $\delta'$ from $B_{t-1}^{+}$ as the base,
%a subset of the coordinates $p$ that will be modified,
%a random value $v_p$ from the refined search space $\Delta$ for each selected coordinate $p$
%and finally add them onto $\delta'$,
%resulting in a new perturbation $\delta''$ which is stored in $B$.

When the search space $\Delta$ has been refined $s$ times,
we get $s$ new perturbations (i.e., set $B$),
resulting in $(2s+k)$ perturbations in the set $B\cup B_{t-1}$.
From them, we choose the smallest-$(s+k)$ perturbations in terms of the dissatisfaction-degree (Line 37).
Algorithm~\ref{alg:1} continues the above procedure on $B_t$ until
an integer adversarial example is found or the number of iterations $T$ is reached.

\medskip
\noindent{\bf Dimensionality reduction.}
Algorithm~\ref{alg:1} depicts the main workflow of our approach which solves the integer adversarial example searching problem by a classification model-based DFO method.
It can be further optimized by a dimensionality reduction technique,
which reduces the search space $\Delta$ into a lower dimensional space, to improve query efficiency.
Dimensionality reduction has been adopted in recent attacks, e.g., AutoZOOM~\cite{TTCLZYC18} and GenAttack~\cite{ASCZHS19}.
Instead of searching in the large search space $\Delta= \mathds{N}_{[-\epsilon,\epsilon]}^{w\times h \times {ch}}$,
we can first search a perturbation $\delta_r$ in a small search space $\Delta_r= \mathds{N}_{[-\epsilon,\epsilon]}^{w_r\times h_r \times {ch}}$
for $w_r\leq w$ and $h_r\leq h$, and scale $\delta_r$ up to $\delta_o$ with the same size as input (i.e. the search space $\Delta$) by applying resizing methods (e.g., bilinear resizing),
resulting in the valid image $\vec{d}\oplus\delta_o$ in the original size. (Please refer to~\cite{TTCLZYC18} and~\cite{ASCZHS19} for the more details of dimensionality reduction.)
By doing so, the query efficiency of our method can be improved while maintaining the attack success rate under
the $\mathds{L}_\infty$ constraint.

\subsection{Illustrative Example}
\begin{figure}[t]
  \centering
  \includegraphics[width=0.95\textwidth]{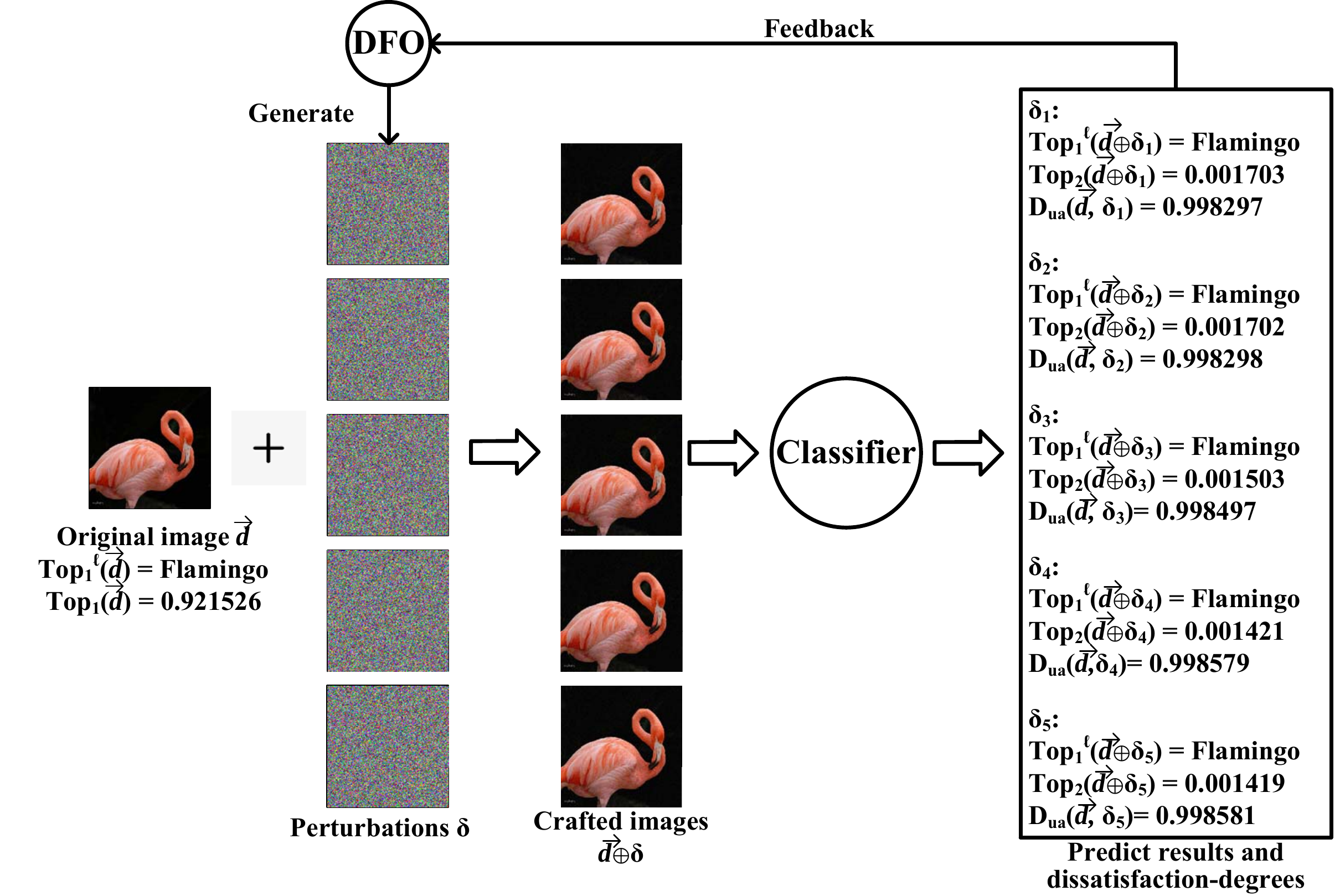}
  \caption{Untargeted attack on a Flamingo image (the first iteration).}
  \label{fig:demo}
\end{figure}

\begin{figure}[t]
  \centering
  \includegraphics[width=0.95\textwidth]{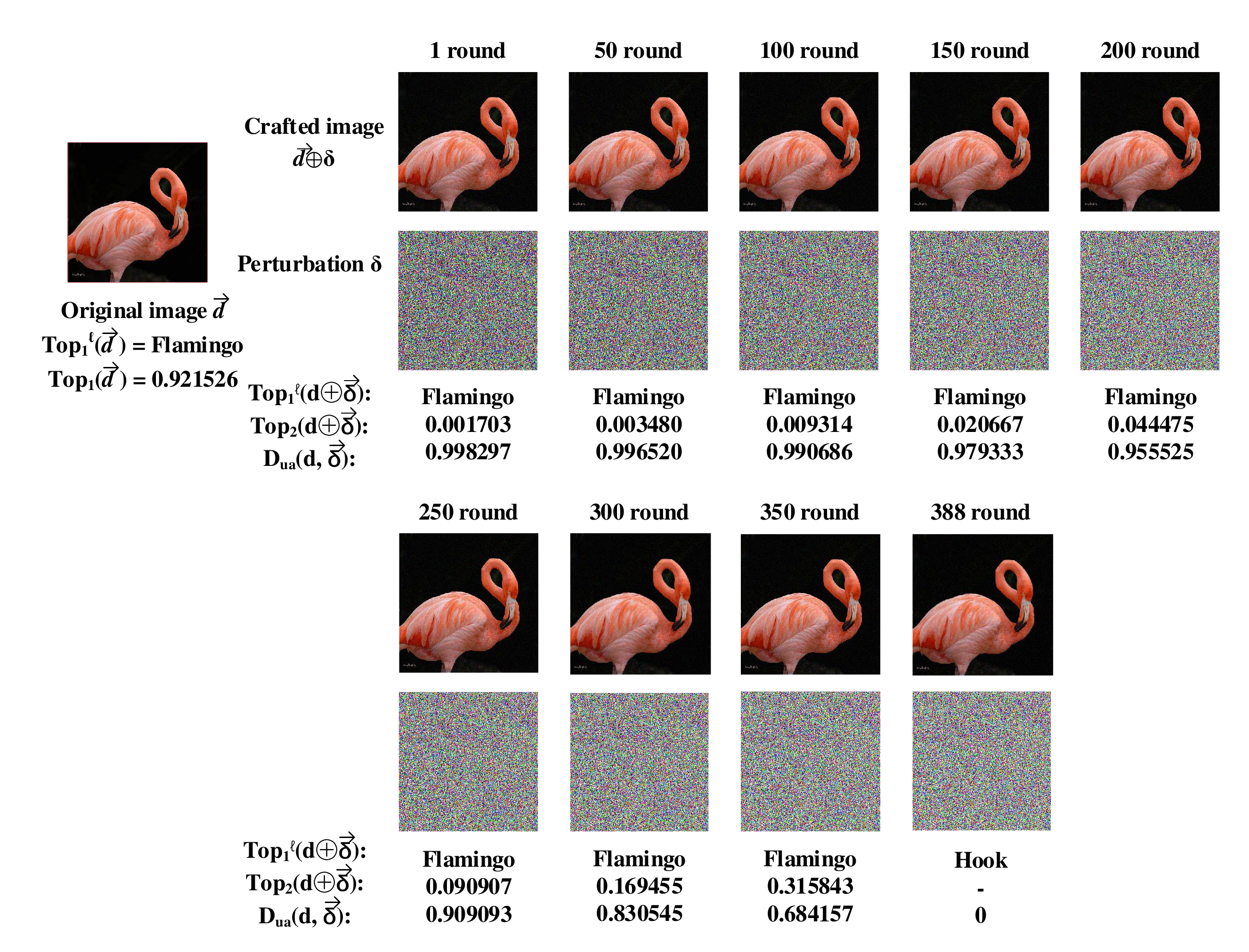}
  \caption{Illustrative example of an untargeted attack on a \emph{Flamingo} image.}\label{fig:demo2}
\end{figure}

We illustrate Algorithm~\ref{alg:1} through an example,
as shown in Figure~\ref{fig:demo}.
The original integer image $\vec{d}$ is an image from the ImageNet dataset and it is classified as the class \emph{flamingo}
by the target classifier Inception-v3.
To launch untargeted attack using this image,
we set the sample size $s$ as $3$ and the ranking threshold $k$ as $2$.
Consider the first iteration,
Algorithm~\ref{alg:1} samples $5$ perturbations $(\delta_i)_{1\leq i\leq 5}$ from $\Delta= \mathds{N}_{[-20,20]}^{w\times h \times {ch}}$
and adds them onto the original image,
resulting in five new images (shown in Figure~\ref{fig:demo}).
Then, it computes the dissatisfaction degrees of these five new images
$(\vec{d}\oplus\delta_i)_{1\leq i\leq 5}$ by
querying the classifier and the dissatisfaction-degree function $D_{\tt ua}$. Among these $5$ perturbations,
$(\vec{d}\oplus\delta_1)$ has the smallest dissatisfaction degree,
hence $\delta_1$ is the best-so-far perturbation.
After more iterations, the results are shown in Figure~\ref{fig:demo2}.
We can see that after the $388$-th iteration, the image with smallest dissatisfaction degree is classified as the class \emph{hook},
but is visually indistinguishable from the original one.

%During the $354$ iterations, the probability of the image falls into the flamingo category is decreased gradually from $0.9057$ at beginning to $0.536$ in the $350$th round?and misclassified into hook in the $354$th round.

%With the execution of iteration, the probability of $cat$, ie $F_{cat}$, is decreasing. The iteration is interrupted when reaches five hundred calls because of that here is another label that its probability is higher than $cat$.

%\begin{center}
%\begin{tabular}{|c|c|c|c|c|}
%\hline
%iter&$L_{predict}$&$P_{1st}$&$P_{2nd}$&$Df$\\
%\hline
%0&\multirow{5}*{tiger cat}&0.3044&0.2236&0.0808\\
%5&~&0.3013&0.2331&0.0682\\
%15&~&0.2783&0.2355&0.0428\\
%25&~&0.2603&0.2388&0.0215\\
%35&~&0.2499&0.2463&0.0036\\
%\hline
%38&tabby&0.2435&0.2410&0\\
%\hline
%\end{tabular}\\
%\textbf{Table 1}~~The probability and the value of $Df$ during the process.\\
%\end{center}

\subsection{Scenario Extension}
%n the previous section, we use the untargeted attack as an example to illustrate how to formulate the adversary example generation problem into an optimization problem, and apply a DFO method to optimize the problem. This %method can be adapted to different problems and scenarios easily.

Our framework is very reflexible and could be potentially adapted to other
scenarios such as:
(1) target classifiers that only output \emph{top-1 class} and \emph{its probability},
and (2) target classifiers that are integrated with defenses,
%attack with different lighting condition,
%and occlusion with tiny black rectangles,
by restricting the search space
or modifying dissatisfaction-degree functions.
%We leave them to future work.

%Our framework is very reflexible and could be adapted to different
%problems, classifiers and scenarios easily such as one-pixel attack,
%attack with different lighting condition,
%occlusion with a single small rectangle,
%occlusion with multiple tiny black rectangles,
%more restricted black-box attacks, and so on.
%These scenario requirements could be easily included into our method by put special restrictions on the search space
%and/or dissatisfaction-degree functions.

For instance, if the adversary only have access to the top-1 class  and its probability,
the dissatisfaction-degree function for untargeted attack can be adapted as follows:
\begin{itemize}%[leftmargin=*]
  \item $D_{\tt ua}^1(\vec{d},\delta):=0$, if $\Ttop_1^{\ell}(\vec{d}\oplus\delta)\neq f_t(\vec{d})$;
  \item $D_{\tt ua}^1(\vec{d},\delta):=\Ttop_1(\vec{d}\oplus\delta)$, otherwise.
\end{itemize}
The dissatisfaction-degree function for targeted attack could be adapted accordingly.
Remark that it is different from label-only attacks in which the adversary has access to
the top-1 class, but \emph{not} its probability.
We leave this to future work.

%% file: experiments.tex
\section{Implementation and Evaluation}
\label{sec:expr}
We %implement our method in a black-box attack tool \tool and also
implement our classification model-based DFO method in \tool based on the framework of RACOS~\cite{yu2016derivative},
for which we implement our new algorithm and manage to engineer to significantly improve its efficiency
and scalability with lots of
domain-specific optimizations. Hereafter, we report experimental results
compared with state-of-the-art white-box and black-box attacks.

\subsection{Dataset \& Setting}

\noindent
\textbf{Dataset.} We use two standard datasets MNIST~\cite{MNIST98}  and ImageNet~\cite{DDSLL009}.
MNIST is a dataset of handwritten digits with $10$ classes (0-9).
%MNIST includes $60000$ images for training and $10000$ images for validation.
We choose the first 200 images out of $10000$ validation images of MNIST as our subjects.

We use the same 100 ImageNet images
as in Section~\ref{sec:Lempstudy}.
(Recall that we randomly choose 100 classes from which
we randomly choose 1 image per class that can be correctly
classified by four classifiers in Keras: ResNet50, Inception-v3,
VGG16 and VGG19.)

\medskip
\noindent
\textbf{Target model.}
For MNIST images, we use a DNN classifier LeNet-1 from the LeNet family~\cite{LBBH98}.
LeNet-1 is a popular target model for MNIST images, e.g.,~\cite{PCYJ17,MJZSXLCSLLZW18,SekhonF19,GuoJZCS18,XieMJXCLZLYS19}.
For ImageNet images, we use  a pre-trained DNN classifier Inception-v3~\cite{SVISW16} which is a  widely used target model for ImageNet images, e.g., \cite{CW17b,CZSYH17,TTCLZYC18,IEAL18,ASCZHS19}.

\begin{table}
\caption{Experiment Settings}\small
\label{tab:set}
\centering
\begin{tabular}{l|l}
  \toprule
  \textbf{Parameter} & \textbf{Setting} \\ \midrule
Max.  $\mathds{L}_\infty$ distance $\epsilon$ & $\epsilon=64$ for MNIST and $\epsilon=10$ for ImageNet. \\ \midrule
 \tabincell{l}{Target class}  &  \tabincell{l}{For MNIST images, the class with 4th largest probability \\is chosen as the target class. \\
  For ImageNet images, the class with 11th largest probability \\is chosen as the target class.} \\ \hline
 \tabincell{l}{Sample size $s$} & \tabincell{l}{{$s=3$ in all the experiments.}} \\ \midrule
\tabincell{l}{Ranking Threshold $k$} &  \tabincell{l}{$k=2$ in all the experiments.} \\ \midrule
\tabincell{l}{Coordinate \\Threshold $u$} & \tabincell{l}{$u=2$ pixels for MNIST images.\\ {$u=10$} pixels for ImageNet images.}\\\hline %The number of coordinates that can be modified in \\ the perturbation during each iteration is about $0.2\%$, \\i.e.,
 Iteration Threshold $T$ & \tabincell{l}{{$T=30000$} in all the experiments.} \\ \midrule
  Timeout Threshold  & \tabincell{l}{3 minutes for MNIST images. \\30 minutes for ImageNet images.} \\ \midrule
  Resized Space $\Delta_r$ & \tabincell{l}{No resize for MNIST images.\\ $100\times100\times3$ for ImageNet images.}\\ \bottomrule
\end{tabular}
\end{table}

\medskip
\noindent
\textbf{Setting.} As shown in Section~\ref{sec:Lempstudy}, the discretization problem can be avoided or alleviated by tuning input parameters for some tools, at the cost of attack efficiency or quality of adversarial examples, except for FGSM and C\&W+GS.
Therefore,  to maximize their TSRs as done in Section~\ref{sec:Lempstudy},
we choose proper step sizes for FGSM
and  enable greedy search for C\&W+GS with parameters recommended by Carlini on MNIST images.
For other tools, we use the parameters
as in their raw papers which are already fine-tuned by the authors for effectiveness and efficiency.
A discussion on turning input parameters refers to Section~\ref{sec:tuning-para}.
Furthermore, some tools only provide implementations for attacking under some specific settings.
If this issue happens, we may not modify their implementations as it may greatly under-estimate their effectiveness
and efficiency, as pointed out by Carlini~\cite{Car19}, hence 
some attacks are not evaluated in all settings. 

%For FGSM, we choose proper parameters (i.e., step sizes) in order to maximize their TSRs as done in Section~\ref{sec:Lempstudy}.
%For C\&W+GS, we enable greedy search
%with the parameters recommended by Carlini on MNIST images\footnote{Remark that C\&W+GS, obtained from Carlini, only implements a%ttacks
%for MNIST images with fine-tuned input parameters.}.

We conduct both untargeted attack and targeted attack %in the experiments,
on a Linux PC running UBUNTU 16.04 LTS with Intel Xeon(R) W-2123 CPU, TITAN Xp COLLECTORS GPU and 64G RAM. Table~\ref{tab:set} lists the other experiment settings.

\begin{table}[t]
  \centering\small
\centering%\setlength{\tabcolsep}{4pt}	
\caption{Results of white-box untargeted attacks}

\label{tab:wu}
\begin{tabular}{|c|c|c|c|c|c|}\hline
{\bf Dataset \& DNN}&{\bf Method}&{\bf SR}&{\bf TSR}&{\bf GAP}\\
\toprule
\multirow{5}*{\tabincell{c}{MNIST \\ LeNet-1}}&FGSM&97\%&97\%&0\%\\
~&BIM&{100\%}&{\bf 100\%}&0\%\\
~&C\&W&{100\%}&88\%&12\%\\
~&C\&W+GS&{100\%}&{\bf 100\%}&0\%\\
~&\tool&{100\%}&{\bf 100\%}&0\%\\
\midrule
%\multirow{4}*{ImageNet}&\multirow{4}*{ResNet50}&FGSM&95\%&95\%&0\%&0.08\\
%~&~&BIM&99\%&{\bf 99\%}&0\%&0.11\\
%~&~&C\&W&99\%&{94\%}&5.05\%&8.56\\
%%~&~&\tool&{\color{red} 96\%}&{\color{red} 96\%}&0\%&{\color{red} 37.2}\\
%~&~&\tool&95\%&95\%&0\%&60.1\\
%\hline
\multirow{4}*{\tabincell{c}{ImageNet\\Inception-v3}}&FGSM&79\%&79\%&0\%\\
~&BIM & {100\%}&{\bf100\%}&0\%\\
~&C\&W&{100\%}&68\%&32\%\\
%~&~&\tool&{\color{red} 95\%}&{\color{red} 95\%}&0\%&{\color{red} 68.3}\\
~&\tool&{99\%}&{\bf 99\%}&0\%\\
\bottomrule
\end{tabular}

\end{table}

\iffalse
\begin{table}[t]
  \centering
\centering%\setlength{\tabcolsep}{4pt}	
\caption{Results of white-box untargeted attacks}
 % \vspace{-3mm}
\label{tab:wu}
\scalebox{1}{
\begin{tabular}{|c|c|c|c|c|c|c|}\hline
{\bf Dataset \& DNN}&{\bf Method}&{\bf SR}&{\bf TSR}&{\bf GAP}&{\bf ATC}\\
\hline
\multirow{5}*{\tabincell{c}{MNIST \\ LeNet-1}}&FGSM&97\%&97\%&0\%&0\\
~&BIM&100\%&{\bf 100\%}&0\%&0.01\\
~&C\&W&100\%&88\%&12\%&0.46\\
~&C\&W+GS&100\%&{\bf 100\%}&0\%&0.47\\
~&\tool&100\%&{\bf 100\%}&0\%&1.9\\
\hline
%\multirow{4}*{ImageNet}&\multirow{4}*{ResNet50}&FGSM&95\%&95\%&0\%&0.08\\
%~&~&BIM&99\%&{\bf 99\%}&0\%&0.11\\
%~&~&C\&W&99\%&{94\%}&5.05\%&8.56\\
%%~&~&\tool&{\color{red} 96\%}&{\color{red} 96\%}&0\%&{\color{red} 37.2}\\
%~&~&\tool&95\%&95\%&0\%&60.1\\
%\hline
\multirow{4}*{\tabincell{c}{ImageNet\\Inception-v3}}&FGSM&79\%&79\%&0\%&0.11\\
~&BIM&100\%&{\bf 100\%}&0\%&0.12\\
~&C\&W&100\%&68\%&32\%&1.41\\
%~&~&\tool&{\color{red} 95\%}&{\color{red} 95\%}&0\%&{\color{red} 68.3}\\
~&\tool&\color{red}{99\%}&{99\%}&0\%&{87}\\
\hline
\end{tabular}}%\vspace{-2mm}
\end{table}
\fi
%\end{center}

\begin{table}[t]
  \centering\small
\centering%\setlength{\tabcolsep}{4pt}	
\caption{Results of white-box targeted attacks}\label{tab:wt}

\begin{tabular}{|c|c|c|c|c|c|}
\toprule
{\bf Dataset \& DNN}&{\bf Method}&{\bf SR}&{\bf TSR}&{\bf GAP}\\ \midrule
\multirow{5}*{\tabincell{c}{MNIST \\ LeNet-1}}& FGSM-1 &84\%&84\% &0\%\\
~&BIM&100\%&{\bf 100\%}&0\%\\
~&C\&W&100\%&75\%&25\%\\
~&C\&W+GS&100\%&{\bf 100}\%&0\%\\
~&\tool&100\%&{\bf 100\%}&0\%\\
\midrule
%\multirow{3}*{ImageNet}&\multirow{3}*{ResNet50}  &FGSM&0\%&-&-&-\\
%~&~&BIM&100\%&{\bf 100\%}&0\%&0.13\\
%~&~&C\&W&99\%&62\%&37.37\%&19.85\\
%%~&~&\tool&{\color{red} \%}&{\color{red} \%}&0\%&{\color{red} }\\
%%~&~&\tool&90\%&{90\%}&0\%&118\\
%~&~&\tool&93\%&{ 93\%}&0\%&182\\
%\hline
\multirow{3}*{\tabincell{c}{ImageNet\\Inception-v3}} &FGSM-1&9\%&9\%&0\%\\
~&BIM&99\%&{\bf99\%}&0\%\\
~&C\&W&100\%&24\%&76\%\\
~&\tool&96\%&{\bf 96\%}&0\%\\
\bottomrule
\end{tabular}
%\begin{tablenotes}
%\item \footnotesize{Note: FGSM does not support ResNet50 and Inception-v3 models.}
%\end{tablenotes}

\end{table}

\iffalse
\begin{table}[t]
  \centering
\centering\setlength{\tabcolsep}{4pt}	
\caption{Results of white-box targeted attacks}\label{tab:wt}

\scalebox{0.9}{
\begin{tabular}{|c|c|c|c|c|c|c|}
\hline
{\bf Dataset \& DNN}&{\bf Method}&{\bf SR}&{\bf TSR}&{\bf GAP}&{\bf ATC}\\ \hline
\multirow{5}*{MNIST}&\multirow{5}*{LeNet-1}&FGSM&3\%&3\%&0\%&0\\
~&~&BIM&100\%&{\bf 100\%}&0\%&0.01\\
~&~&C\&W&100\%&75\%&25\%&1.64\\
~&~&C\&W+GS&100\%&{\bf 100}\%&0\%&1.79\\
~&~&\tool&100\%&{\bf 100\%}&0\%&3.9\\
\hline
%\multirow{3}*{ImageNet}&\multirow{3}*{ResNet50}  &FGSM&0\%&-&-&-\\
%~&~&BIM&100\%&{\bf 100\%}&0\%&0.13\\
%~&~&C\&W&99\%&62\%&37.37\%&19.85\\
%%~&~&\tool&{\color{red} \%}&{\color{red} \%}&0\%&{\color{red} }\\
%%~&~&\tool&90\%&{90\%}&0\%&118\\
%~&~&\tool&93\%&{ 93\%}&0\%&182\\
%\hline
\multirow{3}*{ImageNet}&\multirow{3}*{Inception-v3} &FGSM&0\%&-&-&-\\
~&~&BIM&99\%&{\bf 99\%}&0\%&0.19\\
~&~&C\&W&100\%&24\%&76\%&14.88\\
~&~&\tool&96\%&\color{red}{ 96\%}&0\%&232\\
\hline
\end{tabular}}
%\begin{tablenotes}
%\item \footnotesize{Note: FGSM does not support ResNet50 and Inception-v3 models.}
%\end{tablenotes}
%\vspace{-2mm}
\end{table}
\fi

\subsection{Comparison with White-Box Methods}
%Our method is fast to generate adversarial examples with high success, thus we can compare with white-box method.
Although our method is a black-box one,
we compare the performance with four well-known white-box tools: FGSM, BIM, C\&W
and C\&W+GS, where the implementations are
by their authors.
Since FGSM has nearly no ability to handle targeted attack, we use
one-step target class method (denoted by FGSM-1) of \cite{KurakinGB17}, which can be regarded as the targeted version of FGSM.
%For FGSM and BIM, we tuned the input parameters (i.e., step size)
%in order to maximize their TSRs as in Section~\ref{sec:DPprobleminWild}.
%For C\&W and C\&W+GS, we .
The maximum $\mathds{L}_\infty$ distances are transformed into their maximum ${\bf L}_\infty$ distances accordingly.

The results are shown in Table~\ref{tab:wu} and Table~\ref{tab:wt} for
untargeted and targeted attacks, respectively.
Notice that C\&W+GS only implements attacks for MNIST images,
hence is not applied to ImageNet images.

Overall, our attack \tool achieves close to 100\% attack success rates for both targeted
and untargeted attacks.
In terms of SR,
our tool \tool outperforms FGSM/FGSM-1 and is comparable to the other tools. % FGSM~\cite{GSS15}, BIM~\cite{KGB17} and C\&W~\cite{CW17b}.
In terms of TSR, %BIM outperforms the others, while
\tool is comparable to BIM and
outperforms FGSM, FGSM-1 and C\&W in most cases.
%It is not surprising that our black-box tool \tool requires more time than other white-box methods.

Specifically,
FGSM, FGSM-1, BIM and  C\&W+GS do not have any gap due to
the tuning of step sizes and the greedy search based algorithm.
It is easy to observe that C\&W has a relatively larger gap in targeted attacks on Inception-v3 in ${\bf L}_\infty$ norm setting,
as its TSR is only $24\%$ compared with $100\%$ SR.
Thus, although C\&W outperforms \tool in terms of SR,
\tool outperforms C\&W in most cases in terms of TSR.
We remark that the gap of C\&W is slightly different from
the one given in Table~\ref{tab:TheExpofDP}, as
C\&W-$\textbf{L}_2$ has an input parameter $\kappa$ which can control the confidence.
By increasing $\kappa$, the confidence of real adversarial examples as well as the TSR of C\&W-$\textbf{L}_2$ increase, and the gap can be minimized.
Whereas C\&W-$\textbf{L}_\infty$ does not have this parameter.

\subsection{Comparison with Black-Box Methods}\label{sec:comblack-box}
%As mentioned above, the black-box methods are divided into two categories according to the need to train a new model. One is the substitution model that needs to train a new model, which is similar to the target model in structure and training set. After the new model is trained, apply the white-box method to generate adversarial examples, which may succeed on target model because of the migration aptitude. We chose the substitution model in the experiments of DeepXplore, which won the best paper in SOSP'17.

We compare \tool with well-known
recent black-box methods: substitute model based attacks,
ZOO,  NES-PGD, FD, FD-PSO, and also three concurrent works Bandits, AutoZOOM and GenAttack,
representing all the classes of existing black-box attacks (cf. Section~\ref{sec:related}),  where the implementations are
by their authors.

Recall that it is very difficult
to tune input parameters for those tools without loss of attack efficiency or quality of adversarial examples, hence
we use the parameters as in their raw papers which are already fine-tuned by the authors.
%In order to avoid the side-effect of the selected white-box methods
When evaluating substitute model,
we use FGSM/FGSM-1 and C\&W methods, %respectively.
%For ImageNet, we use ResNet50 as the substitute model for Inception-v3 and vice versa.
and use ResNet50~\cite{HZRS16} as the substitute model for Inception-v3,
the model in ZOO as the substitute model for LeNet-1.
Since ZOO and AutoZOOM use ${\bf L}_2$ distance,
we map our maximum $\mathds{L}_\infty$ distances into maximum ${\bf L}_2$ distances
by considering the worst case of $\mathds{L}_\infty$, namely,
all the pixels are modified by the maximum $\mathds{L}_\infty$ distance.
For instance, the $\mathds{L}_\infty$ distance 10 is approximated by ${\bf L}_2$ distance
$\sqrt{(10/255)^{2}\times(299\times299\times3)}\approx20$
 for $299\times 299\times 3$ images.
Remark that this is not a rigorous mapping, ZOO and AutoZOOM under ${\bf L}_2$ would be easier to find an adversarial example, as
the corresponding ${\bf L}_2$ distances are less restricted.

%The black-box methods we compared including substitution model,  alternative gradient based method (ZOO and NES-PGD), and decision-based method (DBA).
%In order to eliminate the impact of the selected white-box methods when evaluating substitution model, we use both the FGSM and C\&W methods respectively. For ImageNet, we use ResNet50 as the substitution model for Inception-v3 and vice versa. The substitution model for LeNet-1 is from the evaluation of ZOO~\cite{CZSYH17}.
%It is worth to note that, as our tool uses ${\bf L}_\infty$ distance, while DBA and ZOO use ${\bf L}_2$ distance. We map the ${\bf L}_\infty$ limitation used in \tool to the corresponding approximated ${\bf L}_2$ value in the experiments.  Meanwhile, some of the tools are dismissed in the table for certain problems because the corresponding tool has not given the related parameter for certain problem respectively.

The results of untargeted and targeted attacks are given in Table~\ref{tab:bu}  and Table~\ref{tab:bt}.
We can see that our attack \tool achieves close to 100\% attack success rates for both targeted
and untargeted attacks and outperforms all the other tools in terms of TSR no matter targeted or untargeted attacks.
%
%Specifically, %substitute model based attacks can achieve at
%most $38\%$ TSR under untargeted setting,
%and \fu{$3\%$ TSR under targeted setting.}
%
In terms of SR, our tool
is also comparable (or better)
to the other tools. 
One may notice that substitute models perform poorly.
This may due to the difference between training data and architectures 
of the substitute model and the target model,
as the larger gap between the substitute model and the target model is,
the less effective of transferability of adversarial samples is.

%ZOO, NES-PGD, Bandits, AutoZOOM, and GenAttack. %,% and DBA~\cite{RBB17},
%However, all these methods suffer from the discretization problem, they usually have gaps between SR and TSR.
%For instance, the attack success rate of ZOO is dramatically dropped from $89\%$ to $5\%$ on Inception-v3 in untargeted setting.
%Therefore, our tool outperforms the others in terms of TSR.

\begin{table}[t]
  \centering\small
\centering%%\setlength{\tabcolsep}{4pt}
\caption{Results of black-box untargeted attacks, where  FD and FD-PSO do not provide attacks against ImageNet images.
ZOO and AutoZOOM do not provide attacks under  $\mathds{L}_\infty$ distance, so we only compare
our tool with them on ImageNet images.
NES-PGS and Bandits do not provide attacks against MNIST images. Meanwhile, the version of GenAttack's attack against MNIST is buggy}
\label{tab:bu}
\begin{tabular}{|c|c|c|c|c|c|}
\toprule
{\bf Dataset \& DNN}&{\bf Method}&{\bf SR}&{\bf TSR}&{\bf GAP}\\
\midrule
\multirow{5}*{\tabincell{c}{MNIST \\ LeNet-1}}&SModel+C\&W&2.5\%&2.5\%&0\%\\
~&SModel+FGSM&20\%&20\%&0\%\\
%~&~&DBA&99.5\%&84\%&15.58\%&11.89\\
%~&~&ZOO&\textbf{100\%}&96.5\%&3.5\%\\
~&FD&94.5\%&94.5\%&0\%\\
~&FD-PSO&46.5\%&46.5\%&0\%\\
~&\tool&{100\%}&{\bf 100\%}&0\%\\
\midrule
%\multirow{4}*{ImageNet}&\multirow{4}*{ResNet50}&
%SModel+C\&W&2\%&1\%&50\%&1.29\\
%~&~&SModel+FGSM&24\%&24\%&0\%&0.12\\
%%
%%SModel+C\&W&0\%&-&-&-\\
%%~&~&SModel+FGSM&0\%&-&-&-\\
%%~&~&DBA&100\%&44\%&56\%&137.79\\
%~&~&Bandits&94\%&21\%&77.7\%&22.7\\
%%~&~&\tool&97\%&{\bf 97\%}&0\%&150.66\\
%~&~&\tool&95\%&{\bf 95\%}&0\%&60.1\\
%%~&~&\tool&{\color{red} 96\%}&{\bf 96\%}&0\%&{\color{red} 37.2}\\
%\hline
\multirow{8}*{\tabincell{c}{ImageNet\\Inception-v3}}&
%SModel+C\&W&0\%&-&-&-\\
%~&~&SModel+FGSM&0\%&-&-&-\\
SModel+C\&W&6\%&6\%&0\%\\
~&SModel+FGSM&38\%&38\%&0\%\\
%~&~&Decision-Based&-&-&-&-\\
%~&~&DBA&100\%&34\%&66\%&254.2\\
%~&~&ZOO&73\%&3\%&95.79\%&113\\
~&ZOO&89\%&5\%&94.3\%\\
~&AutoZOOM&{100\%}&57\%&43\%\\
~&NES-PGD&{100\%}&77\%&23\%\\
~&Bandits&100\%&12\%&88\%\\
~&{ GenAttack}&{{100\%}}&{93\%}&{ 7\%}\\
~&\tool&{ 99\%}&{\bf 99}\%&{0\%}\\
%~&~&\tool&{\color{red} 95\%}&{\bf 95\%}&0\%&{\color{red} 68.3}\\
\bottomrule
\end{tabular}
\end{table}

\begin{table}[t]
  \centering\small
\centering%\setlength{\tabcolsep}{5pt}
\caption{Results of black-box targeted attacks, where Bandits does not support targeted attack}
\label{tab:bt}
\begin{tabular}{|c|c|c|c|c|c|}
\toprule
{\bf Dataset \& DNN}&{\bf Method}&{\bf SR}&{\bf TSR}&{\bf GAP}\\
\midrule
\multirow{5}*{\tabincell{c}{MNIST\\LeNet-1}}&SModel+C\&W&1.5\%&1.5\%&0\%\\
~&SModel+FGSM-1&5\%&5\%&0\%\\
%~&~&DBA&95.5\%&79\%&17.28\%&12.17\\
%~&~&ZOO&\textbf{100\%}&97.5\%&2.5\%\\
~&FD&72\%&72\%&0\%\\
~&FD-PSO&6.5\%&6.5\%&0\%\\
~&\tool&{100}\%&{\bf 100\%}&0\%\\
\midrule
%\multirow{3}*{ImageNet}&\multirow{3}*{ResNet50}&
%SModel+C\&W&2\%&2\%&0\%&27.21\\
%~&~&SModel+FGSM&0\%&-&-&-\\
%%~&~&Bandits&-&-&-&-\\
%%~&~&DBA&95\%&21\%&77.89\%&335\\
%%~&~&\tool&{\color{red} 95\%}&{\color{red} 95\%}&0\%&{\color{red} 109}\\
%%~&~&\tool&87\%&{\bf 87\%}&0\%&211.32\\
%~&~&\tool&93\%&{\bf 93\%}&0\%&182\\
%\hline
\multirow{8}*{\tabincell{c}{ImageNet\\Inception-v3}}&
SModel+C\&W&1\%&1\%&0\%\\
~&SModel+FGSM-1&2\%&2\%&0\%\\
%~&~&DBA&87\%&25\%&71.3\%&544\\
%~&~&ZOO&62\%&6\%&90.32\%&716.04\\
~&ZOO&69\%&5\%&92.7\%\\
~&AutoZOOM&95\%&43\%&54.7\%\\
~&NES-PGD&{100\%}&47\%&53\%\\
%~&Bandits&-&-&-\\
~&{GenAttack}&{100\%}&{84\%}&{16\%}\\
%~&~&\tool&{\color{red} 92\%}&{\color{red} 92\%}&0\%&{\color{red} 309.2}\\
~&\tool&{96\%}&{\bf 96\%}&0\%\\
\bottomrule
\end{tabular}
\end{table}

\begin{table}[t]
  \centering\small
  \centering%\setlength{\tabcolsep}{2pt}
\caption{Comparison with average query times and corresponding MSE, where  the queries of our tool is computed on integer adversarial examples,
while it is computed on real adversarial examples for the others. }\label{tab:query}

\begin{tabular}{|c|c|c|c|c|c|c|}
\toprule
\multirow{2}*{\bf Dataset \& DNN}&\multirow{2}*{\bf Method}&\multicolumn{2}{c|}{\bf Untargeted}&\multicolumn{2}{c|}{\bf Targeted}\\\cline{3-6}
%\hline
~&~&Query&MSE&Query&MSE\\
\midrule
%\multirow{6}*{ImageNet}&\multirow{6}*{Inception-v3}&NES-PGD&4741&{$8.4\times{10^{-4}}$}&13421&{$9.0\times {10^{-4}}$}\\
%~&~&{Bandits}&2923&{$1.4\times {10^{-3}}$}&-&{-}\\
%~&~&{AutoZOOM (${\bf L}_2=20$)}&\textbf{2224}&{$9.2\times{10^{-4}}$}&14322&{$1.2\times {10^{-3}}$}\\
%~&~&{AutoZOOM (${\bf L}_2=12$)}&{4971}&{$4.6\times{10^{-4}}$}&21356&{$4.9\times {10^{-4}}$}\\
%~&~&genattack&{4008}&{$6.3\times{10^{-4}}$}&{12369}&{$9.2\times{10^{-4}}$}\\
%~&~&\tool&\color{red}{4746}&{$2.6\times{10^{-4}}$}&\color{red}{\textbf{12740}}&{$3.4\times{10^{-4}}$}\\
\multirow{3}*{\tabincell{c}{MNIST\\LeNet-1}}
&FD&1568&3.4e-2&\textbf{1568} & 3.5e-2\\
~&{FD-PSO}&10000&\textbf{2.5e-2}&10000 & \textbf{2.5e-2}\\
%~&~&{ZOO}&126885&2.4e-5&187763&3.0e-5\\
~&\tool&\textbf{817}&\textbf{3.7e-2}&{\bf 1593}&\textbf{5.1e-2}\\
\midrule
\multirow{6}*{\tabincell{c}{ImageNet\\Inception-v3}}&ZOO&85368&\textbf{1.7e-5}&203683&\textbf{3.6e-5}\\
~&AutoZOOM&\textbf{2224} & 9.2e-4 & 14322 &1.2e-3\\
%~&~&{AutoZOOM (${\bf L}_2=12$)}& 4971& 4.6e-4& 21356 &4.9e-4\\
%~&{AutoZOOM (${\bf L}_2=15$)}& 3816& 6.7e-4& 22369 &6.9e-4\\
~&{NES-PGD} & 4741 & 8.4e-4& 13421 & 9.0e-4\\
~&{Bandits} & 4595 & 1.4e-3 & -  & - \\
~&GenAttack&{4008}&6.3e-4& {\textbf{12369}} & 9.2e-4\\
~&\tool&{\bf 4746}&{\textbf{2.6e-4}}&{\textbf{12740}}&{\textbf{3.4e-4}}\\
\bottomrule
\end{tabular}
\end{table}

\medskip
%\noindent\textbf{Comparing query times.} %and GenAttack~\cite{ASCZHS19}.}
\subsection{Query Comparison.}
In many black-box scenarios, the attacker has a limited number of queries to the classifier.
Therefore, we report the average number of queries of the black-box attacks
in Table~\ref{tab:query}, where substitute model based attack  is excluded
due to its low SR.
We remark that ZOO is regarded as baseline,
the others are state-of-the-art query-efficient tools.

%We compare query efficiency using Inception-v3 as the target classifier which is
%widely used for adversarial attack.

%ZOO and NES-PGD are gradient estimation based attacks that respectively leverage
%C\&W and PGD attack methods.
%AutoZOOM is an attack tool that uses autoencoder to enhance query efficiency of ZOO.
%Bandits is an optimized version of NES-PGD by using bandit optimization.
%GenAttack is a genetic algorithm based attack tool without gradient estimation. % as well as dimensionality reduction.
%natural evolution strategy
%
%NES-PGD is a natural evolutionary-based tool.
%
%Genetic algorithms are population-based
%gradient-free optimization strategies, previously used to find
%find adversarial examples to fool pdf malware classifiers~\cite{xu2016automatically}.
%We remark that all these tools are black-box attack tools which focus on query-limited scenarios.

%As aforementioned, Bandits does not support targeted attack, therefore we only conduct untargeted attacks for Bandits.

\medskip\noindent
{\bf On attack against Inception-v3}, our tool \tool
outperforms all the other tools for targeted attacks, except for GenAttack,
which is slightly better than \tool.
For untargeted attacks, our tool \tool outperforms the baseline tool ZOO
and comparable to other tools.
Recall that our tool \tool outperforms all these tools in terms of TSR.

We remark that ZOO and AutoZOOM are tested under the ${\bf L}_2$ distance 20, which is less restricted than the $\mathds{L}_\infty$ distance 10 used for the other tools.
Indeed, in untargeted attack setting,
the average ${\bf L}_2$ distance of %of adversarial examples crafted by
our tool is 8.33.
Whereas the average query times of AutoZOOM becomes 4971 (worse than ours)
if ${\bf L}_2=12$.

\medskip\noindent
{\bf On attack against LeNet-1},
 our tool \tool outperforms both of them in almost all cases,
exception that FD uses less query times than \tool for targeted attacks.
Note that our tool \tool achieves higher attack success attack rate than FD and FD-PSO in terms of
both SR/TSR.
One may notice that
the query times of FD and FD-PSO are same between untargeted and targeted attacks.
This is due to the implementations of FD and FD-PSO (confirmed by some authors of \cite{BHLS18}).

Furthermore, we also report the average Mean Square
Error (MSE)  of the adversarial examples in Table~\ref{tab:query}.
We can observe that our tool \tool outperforms most of the other tools on attacks against Inception-v3. FD and FD-PSO are slightly better than \tool on attacks against LeNet-1, at the same order of magnitude. ZOO outperforms all the other tools in terms of MSE against Inception-v3, but at the cost of huge number of query times. % as we can see from the data.

\subsection{Attack Classifiers with Defense}
To show the effectiveness of our approach,
we use our tool to attack the HGD defense~\cite{LLDPH018}, which won the first place of
NIPS 2017 competition on defense against adversarial attacks. HGD defense is a typical denoising based defensing methods for image classification.
The whole classification system is an ensemble of 4 independent models and their denoiser (ResNet, ResNext, InceptionV3, inceptionResNetV2). We conduct untargeted attacks against this model
using the same 100 ImageNet images and parameters as previously,
exception that the $\mathds{L}_\infty$ distance $\epsilon$ is $32$ according to the NIPS 2017 competition.
% (note that it is different  from the distance used in their raw paper).
%The coordinate threshold is set to $200$.
Our tool achieves $100\%$ TSR in the experiments, indicating the effectiveness of  \tool.
This benefits from the advantage of our classification model-based derivative-free
optimization method, which does not rely on the gradient of the objective function, but instead,
learns from samples of the search space, hence suitable for attack systems that are non-differentiable 
or even unknown but only testable.

MNIST Adversarial Examples Challenge~\cite{mnistchallenge19}
is another widely recognized attack problem. It uses adversarial training for defensing.
We use the same 200 MNIST images as previously on the attack of this problem.
Our tool \tool achieves $10.5\%$ TSR, the same as the current best white-box attack ``interval attacks'', which is publicly reported on the webpage of the challenge.
The images on which the attacks succeed by both methods are exactly same, and the time costs of both tools are also similar.

%% file: discussNormalization.tex
\subsection{Discussion on Normalization}\label{sec:discnorm}
In this section, we first survey several typical normalization
and then discuss why it is non-trivial to infer the normalization in black-box scenario.

Computer vision usually require some normalization (also known as preprocessing)  because the original
input comes in a form that is difficult for many deep learning architectures to
represent. The images should be normalized so that their pixels all lie in the same,
reasonable range, like $[0,1]$ or $[-1, 1]$ or $[-0.5,0.5]$.
Mixing images that lie different ranges will usually result in failure~\cite{GoodfellowBC16}.

Dataset augmentation is a kind of preprocessing for the training set
only. Other kinds of normalization are applied to both the train and the test sets with
the goal of putting each example into a more canonical form in order to reduce the
amount of variation that the model needs to account for.
Therefore, in this section, we only discuss normalization for test set. We do not consider
size-normalization, i.e., normalizing images into same size, which would make
inference more difficult.
Specifically, we will present several widely used approaches for normalizing images into some range.

\subsubsection{Normalization for MNIST Images}
The original MNIST images are black and white images and the resulting images are grayscale images as a result of the anti-aliasing technique used by the normalization algorithm.
Each pixel of a grayscale image has only one channel whose value is an integer ranging from 0 to 255, where 0 means background (white), 255 means foreground (black).
For classification of handwritten digits, the value of each pixel is usually normalized into the range $[0,1]$
by dividing $255$.

\subsubsection{Normalization for ImageNet Images}
The ImageNet images are represented in red, green and blue colors,
then, each pixel of an ImageNet image has three channels.
The value of each channel is an integer ranging from 0 to 255.
We list three different normalization below.

\begin{itemize}%[noitemsep,topsep=0pt,leftmargin=*]
  \item {\bf The Inception-v3 model in ZOO~\cite{CZSYH17}:} the integer value $i$ of each channel is normalized into a real value $r$ as follows:
  \[r=\frac{i}{255}-0.5.\]   Obviously, the range of $r$ is $[-0.5,0.5]$.

  \item {\bf The Inception-v3 model in Keras~\cite{keras19}:}  the integer value $i$ of each channel is normalized into a real value $r$ as follows:
  \[r=\frac{2\times i}{255}-1.\] Obviously, the range of $r$ is $[-1,1]$.

  \item {\bf The VGG and ResNet models in Keras~\cite{keras19}:} the integer value $i$ of each channel $j$ ($j=1,2,3$) is normalized into a real value $r$ as follows:
  \[r=i-{\tt mean}_j \]
  where ${\tt mean}_j=\frac{\sum_{d\in {\tt TDS}} \sum_{(w,h,j)\in P} d[w,h,j]}{\frac{|P|}{3}\times |{\tt TDS}|}$, and ${\tt TDS}$ denotes the training dataset.
  Recall that $P$ denotes the set of coordinates.
     Obviously, the range of $r$ is $[-255,255]$ and depends on the training dataset.

  \item {\bf The DenseNet model in ~\cite{HLMW17} and ResNet model in torch~\footnote{https://github.com/facebook/fb.resnet.torch}}: the integer value $i$ of each channel $j$ is normalized into a real value $r$ as follows:
  \[r=\frac{i-{\tt mean}_j}{{\tt std}_j}\]
  where ${\tt mean}_j$ is defined the same as above, the standard deviation ${\tt std}_j$ is defined as follows:
  \[{\tt std}_j=\sqrt{\frac{\sum_{d\in {\tt TDS}} \sum_{(w,h,j)\in P} (d[w,h,j]-{\tt mean}_j)^2}{\frac{|P|}{3}\times |{\tt TDS}|}}.\]
     Obviously, the range of $r$ is $[-\infty,\infty]$ and depends on the training dataset.
\end{itemize}

\subsubsection{Normalization for CIFAR-10 Images}

\begin{itemize}%[noitemsep,topsep=0pt,leftmargin=*]
 \item {\bf The DenseNet model in  ~\cite{HLMW17}}: the integer value $i$ of each channel $j$ is normalized into a real value $r$ as follows:
  \[r=\frac{i-{\tt mean}_j}{{\tt std}_j}\]
  where ${\tt mean}_j$ and ${\tt std}_j$ are defined the same as above.
  Then, the range of $r$ is $[-\infty,\infty]$ and depends on the training dataset.

 \item {\bf The maxout model in~\cite{GoodfellowWMCB13} (cf. \cite[Page 472]{GoodfellowBC16})}:
 the integer value $i$ of each channel $j$ is normalized into a real value $r$ as follows:
  \[r=s\times \frac{i-{\tt mean}_j}{\max\{10^{-8},{\tt std}_j\}}\]
  where ${\tt mean}_j$ and ${\tt std}_j$ are defined the same as above.
 The extremely low value $10^{-8}$ is introduced to avoid division by 0.
 The scale parameter $s$ is chosen to make each individual pixel have standard deviation across examples
close to 1. The range of $r$ depends on $s$ and dataset.

 \item {\bf The model in~\cite{CNL11} (cf. \cite[Page 472]{GoodfellowBC16})}:
 the integer value $i$ of each channel $j$ is normalized into a real value $r$ as follows:
  \[r=\frac{i-{\tt mean}_j}{{\tt std}_j'}\]
  where ${\tt mean}_j$ is defined the same as above and ${\tt std}_j'$ is defined as follows:
  \[{\tt std}_j'=\sqrt{10+\frac{\sum_{d\in {\tt TDS}} \sum_{(w,h,j)\in P} (d[w,h,j]-{\tt mean}_j)^2}{\frac{|P|}{3}\times |{\tt TDS}|}}.\]

The range of $r$ is $[-\frac{255}{\sqrt{10}},\frac{255}{\sqrt{10}}]$ and depends on the training dataset.
Note that $10$ is introduced to avoid division by 0

\end{itemize}

\subsubsection{Discussion}
As shown above, we can observe that:
(1) the same network model on the same dataset (e.g., Inception-v3 and ResNet models on ImageNet images) in different tools  may use different normalization;
(2) the same tool may use different normalization for different models even on the same dataset (e.g.,  Inception-v3 and ResNet in Keras on ImageNet);
and (3) there are several kinds of normalization and the parameters of normalization may depend on training dataset.
Once the normalization is known, its corresponding denormalization can be implemented by choosing a
rounding mechanism, e.g., round up, round down, round to the nearest integer.
In black-box scenario, the adversary can only query discrete integer images to the oracle classifier and get the output, without the knowledge
of training dataset, architecture, parameters and normalization of the classifier.
To our knowledge, in general, it is non-trivial to infer how the normalization is implemented by a classifier.

%% file: discussParameter.tex
\subsection{Results of Parameter Tuning}
\label{sec:app1}
% The results of parameter tuning are shown in Tables ~\ref{tab:Appendix1}, \ref{tab:Appendix2} and \ref{tab:Appendix3}.
%The bold line represents the parameters which been shown before and the first line is the default parameters provided by the author or third-party toolkits.

% For JSMA, we tune the input parameter $\theta$, as
% $\theta \times b$ is the perturbation step size,
% where $b$ denotes the upper bound of pixel values.
% Increasing the value of $\theta$ can improve TSRs.

% For L-BFGS, we tune the input parameter $\epsilon$ which relaxes the distance constraint.
% Furthermore, we also tune an input parameter
% provided by FoolBox built-in criteria
% which controls the confidences of adversarial examples.
% Both of them can be used to improve TSRs.

%the first direction of fine-tuning is to increase the $\epsilon$,
%which could lessen the distance constraint,
%then find more obvious perturbations to alleviate discretization problem,
%the second method is to generate more strong examples
%by FoolBox built-in criteria,
%it will find stronger adversarial examples until its confidence higher than user's inputs.

Strategy S1 aims at controlling perturbation step sizes.
Among 9 tools, NES-PGD, DBA, Bandits and GenAttack provide such adaptive mechanism.
%that can be tuned by input parameters,
We limit the $L_\infty$ distance at discrete domain firstly,
then constraint the perturbation step sizes as integer numbers.
e.g., the parameters $\alpha$ in NES-PGD, `step adaption' in DBA and `adaptive' in GenAttack.
Using this strategy, the TSR of NES-PGD and DBA increases,
but the TSR of Bandits and GenAttack does not increases (cf. Section~\ref{sec:theory_study} for reasons).
The results are shown in Table~\ref{tab:tunepara1}.
We can observe that the TSR of NES-PGD increases at the cost of higher query times,
due to the fact that limiting of the dynamic adjustment of step size or learning rate,
will affect the attack efficiency of this method.

\begin{table}[htbp]\small
\caption{Parameters and results of S1}
\centering
\label{tab:tunepara1}
\begin{tabular}{cccccc}
\multicolumn{6}{l}{NES-PGD~\cite{IEAL18} attack on ImageNet+InceptionV3} \\
\multicolumn{6}{l}{Version1 (default): $\epsilon = 0.05, \alpha \in [1e-2, 5e-5]$} \\
\multicolumn{6}{l}{Version2: $\epsilon = 10/255, \alpha \in [1e-2, 5e-5]$} \\
\multicolumn{6}{l}{Version3: $\epsilon = 10/255, \alpha = 1/255$} \\
\toprule
Version  & SR & TSR & MSE  & Discr. Error & Avg. Queries \\
\midrule
1  & 100\% & 53\%  & 8.77e-04 & $\approx 0.5$ & 12470  \\
2  & 100\% & 47\%  & 5.46e-04 & $\approx 0.5$ & 16168  \\
3  & 74\%  & 73\%  & 5.14e-04 & 0.16  & 214552 \\
\bottomrule
\\

\multicolumn{6}{l}{DBA~\cite{BRB18} attack on ImageNet+VGG19} \\
% \multicolumn{6}{l}{Version1 (default): iteration = 5000, $\alpha = 0.01$, step adaptation = 1.5}\\
\multicolumn{6}{l}{Version1 (baseline): iteration = 1000, $\alpha = 0.01$, step adaptation = 1.5} \\
\multicolumn{6}{l}{Version2: iteration = 1000, $\alpha = 1/255$, step adaptation = 1}\\
\toprule
Version  & SR & TSR & MSE & Discr. Error  \\ % ATC \\
\midrule
% 1  & 100\%  & 28\%  & 1.91e-05   & $\approx 0.5$  \\ %1755.91 \\
1  & 100\%  & 51\%  & 3.56e-04 & $\approx 0.5$  \\ % 401.06 \\
2  & 100\%  & 62\%  & 0.016    & $\approx 0.5$  \\ %587.08 \\
\bottomrule\\

\multicolumn{6}{l}{Bandits~\cite{IEM2018PriorCB} attack on ImageNet+InceptionV3} \\
\multicolumn{6}{l}{Version1 (default): $\epsilon$ = 0.05, $\alpha = 0.01$} \\
\multicolumn{6}{l}{Version2: $\epsilon$ = 10/255, $\alpha = 0.01$ } \\
\multicolumn{6}{l}{Version3: $\epsilon$ = 10/255, $\alpha = 1/255$} \\
\toprule
Version  & SR & TSR & MSE    & Discr. Error  & Avg. Queries \\
\midrule
1  & 94\%  & 11\% & 2.08e-03 & $\approx 0.5$  & 2715 \\
2  & 92\%  & 12\% & 1.4e-03  & $\approx 0.5$  & 2923 \\
3  & 92\%  & 11\% & 1.23e-03 & $\approx 0.5$  & 2950 \\
\bottomrule\\

\multicolumn{6}{l}{GenAttack~\cite{ASCZHS19} attack on ImageNet+InceptionV3} \\
\multicolumn{6}{l}{Version1 (default): $\epsilon = 0.05, \alpha \approx 0.15$, adaptive=True} \\
\multicolumn{6}{l}{Version2: $\epsilon = 10/255, \alpha \approx 0.1$, adaptive=True} \\
\multicolumn{6}{l}{Version3: $\epsilon = 10/255, \alpha = 0.1$, adaptive=False} \\
\toprule
Version  & SR & TSR & MSE & Discr. Error & Avg. Queries \\
\midrule
1  & 100\% & 91\% & 1.61e-03 & $\approx 0.5$   & 24728 \\
2  & 97\%  & 56\% & 1.05e-03 & $\approx 0.5$   & 33273 \\
3  & 99\%  & 46\% & 4.39e-04 & $\approx 0.5$   & 45576 \\
\bottomrule
\end{tabular}
\end{table}
Strategy S2 aims at minimizing the ratio of discretization error against the overall perturbations.
L-BFGS, DeepFool, DeepXplore and DBA
provide input parameters related to this strategy.
For L-BFGS, we increase `initial const', which could relax the distance constraint,
and reduce the binary search which could also avoid constraint enhancement.
For DeepFool, `overshot' controls the distance cross polyhedral boundary,
as same as DeepXplore, `step' multiplies with gradient to get the perturbation.
For DBA, since adversarial examples are starting from target images, if we reduce `iteration', the distance between adversarial and original image will be reduced.
Experimental results show that the TSR of DBA
can increase at the cost of higher overall perturbations, but
the TSR of L-BFGS, DeepFool and DeepXplore does not increase obviously.

\begin{table}[t]
\caption{Parameters and results of the S2}
\label{tab:tunepara2}
\centering \small
\begin{tabular}{ccccc}

\multicolumn{5}{l}{L-BFGS~\cite{SZSBEGF14} attack on ImageNet+Inception-v3}\\
\multicolumn{5}{l}{Version1 (default): initial const = 1e-2, binary search steps = 5} \\
\multicolumn{5}{l}{Version2: initial const = 1e-2, binary search steps = 3} \\
\multicolumn{5}{l}{Version3: initial const = 5e-2, binary search steps = 3} \\
\toprule
Version  & SR & TSR & MSE  & Discr. Error \\
\midrule
1  & 100\% &77\% &2.27e-05 & $\approx 0.5$ \\
2  & 100\% &76\% &2.29e-05 & $\approx 0.5$ \\
3  & 100\% &79\% &4.29e-05 & $\approx 0.5$ \\
\bottomrule \\

\multicolumn{5}{l}{DeepFool~\cite{MFF16} attack on ImageNet+ResNet34} \\ \toprule
Overshoot   & SR & TSR & MSE  & Discr. Error  \\ % & $L_\infty$-distance & ATC
\midrule
1.02  & 100\% &23\% &2.27e-05 & $\approx 0.5$ \\ % & 21.78       & 0.67
1.5   & 100\% &23\% &3.79e-05 & $\approx 0.5$ \\ % & 16.86 & 0.34
2     & 100\% &24\% &6.16e-05 & $\approx 0.5$ \\ % & 23.13 & 0.3
\bottomrule
\\
\multicolumn{5}{l}{DeepXplore~\cite{PCYJ17} attack on ImageNet+ResNet50\&VGG16\&19} \\
\multicolumn{5}{l}{Version1 (default): weight diff=1, step=10} \\
\multicolumn{5}{l}{Version2: weight diff=1, step=20} \\
\multicolumn{5}{l}{Version3: weight diff=2, step=10} \\
\toprule
Version  & SR & TSR & MSE & Discr. Error  \\ % & $L_\infty$-distance & ATC
\midrule
1  & 65\% &28\% &2.14e-02 & $\approx 0.5$ \\ % & 255 & 56.12
2  & 64\% &25\% &2.25e-02 & $\approx 0.5$ \\ % & 255 & 68.05
3  & 65\% &29\% &2.22e-02 & $\approx 0.5$ \\ % & 255 & 187.86
\bottomrule
\\
\multicolumn{5}{l}{DBA~\cite{BRB18} attack on ImageNet+VGG19} \\
\multicolumn{5}{l}{Version1 (default): iteration = 5000, $\alpha = 0.01$, step adaptation = 1.5}\\
\multicolumn{5}{l}{Version2: iteration = 1000, $\alpha = 0.01$, step adaptation = 1.5} \\
% \multicolumn{5}{l}{Version2: iteration = 1000, $\alpha = 1/255$, step adaptation = 1} \\
\toprule
Version  & SR & TSR & MSE      & Discr. Error     \\ \midrule% ATC \\ 
1  & 100\%  & 28\%  & 1.91e-05 & $\approx 0.5$    \\ %1755.91 \\
2  & 100\%  & 51\%  & 3.56e-04 & $\approx 0.5$    \\ %401.06 \\
% 3  & 100\%  & 62\%  & 0.016    & $\approx 0.5$  \\ %587.08 \\
\bottomrule
% $\epsilon$ & SR & TSR & MSE & $L_\infty$-distance\\ % & ATC\\
% \hline
% {1e-05}  & 100\% &72\% &2.78e-06 & 18.89\\ % & 52.72\\
% 1e-04 	 & 100\% &95\% &4.85e-06 & 17.10\\ % & 49.06\\
% 1e-03	 & 100\% &98\% &7.72e-06 & 33.78 \\ %& 37.30\\
% \hline
% Confidence & SR & TSR & MSE & $L_\infty$-distance \\ %& ATC\\
% \hline
% 0        & 100\% &72\% &2.78e-06 & 18.89 \\ %& 52.72\\
% 0.5 	   & 100\% &81\% &5.15e-06 & 28.37 \\ %& 76.37\\
% 0.75     & 100\% &94\% &8.33e-06 & 25.71 \\ %& 78.81\\
%\multicolumn{6}{l}{C\&W-$\textbf{L}_2$~\cite{CW17b} attack on ImageNet+InceptionV3} \\
%\hline
%$\kappa$  & SR & TSR 	   & MSE 	& $L_\infty$-distance & ATC\\
%\hline
%0         & 100\% &10\%    &1.51e-06 & 11.77      & 541.31 \\
%15        & 100\% &99\%    &5.88e-06 & 24.54      & 570.79 \\
%30        & 100\% &100\%   &1.27e-05 & 39.22      & 623.40 \\
%\hline
%\\
\end{tabular}%\vspace{-2mm}
\end{table}

Strategy S3 aims at enhancing the robustness of real adversarial samples against the discretization problem.
Both C\&W and ZOO provide such input parameters
(i.e., $\kappa$). The results are shown in Table~\ref{tab:tunepara3}.
We can observe that the TSR of C\&W increase at the cost of higher overall perturbations (cf. mean-square error (MSE) in Table~\ref{tab:tunepara3}).
However, it does not work for ZOO, although ZOO is a black box version of C\&W
by leveraging gradient estimation.
This is because that ZOO fails to find adversarial samples when the confidence
constraint $\kappa$ increasing.

\begin{table}[h]\small
\caption{Parameters and results of the S3}
\label{tab:tunepara3}
\centering
\begin{tabular}{ccccc}
% $\epsilon$ & SR & TSR & MSE & $L_\infty$-distance\\ % & ATC\\
% \hline
% {1e-05}  & 100\% &72\% &2.78e-06 & 18.89\\ % & 52.72\\
% 1e-04 	 & 100\% &95\% &4.85e-06 & 17.10\\ % & 49.06\\
% 1e-03	 & 100\% &98\% &7.72e-06 & 33.78 \\ %& 37.30\\
% \hline
% Confidence & SR & TSR & MSE & $L_\infty$-distance \\ %& ATC\\
% \hline
% 0        & 100\% &72\% &2.78e-06 & 18.89 \\ %& 52.72\\
% 0.5 	   & 100\% &81\% &5.15e-06 & 28.37 \\ %& 76.37\\
% 0.75     & 100\% &94\% &8.33e-06 & 25.71 \\ %& 78.81\\
\multicolumn{5}{l}{C\&W-$\textbf{L}_2$~\cite{CW17b} attack on ImageNet+InceptionV3} \\
\toprule
$\kappa$  & SR & TSR 	   & MSE 	  & Discr. Error \\ % & $L_\infty$-distance & ATC\\
\midrule
0         & 100\% &10\%    &1.51e-06  & $\approx 0.5$       \\ % & 11.77 & 541.31 \\
15        & 100\% &99\%    &5.88e-06  & $\approx 0.5$       \\ % & 24.54 & 570.79 \\
30        & 100\% &100\%   &1.27e-05  & $\approx 0.5$       \\ % & 39.22 & 623.40 \\
\bottomrule
\\
\multicolumn{5}{l}{ZOO~\cite{CZSYH17} attack on ImageNet+Inception-v3} \\
% \multicolumn{6}{l}{Version1 (default): init const=10, $\kappa=0$} \\
% \multicolumn{6}{l}{Version2: init const=20} \\
% \multicolumn{6}{l}{Version3: $\kappa=15$} \\
\toprule
$\kappa$  & SR & TSR & MSE & Discr. Error \\ % & $L_\infty$-distance & ATC \\
\midrule
0   & 58\% & 6\% & 1.07e-05 & $\approx 0.5$ \\ % & 9.29  & 895.49 \\
15  & 0\% & - & - & -  \\ % 59\% & 5\% & 1.14e-05 & 8.65  & 908.59 \\
30  & 0\% & - & - & -  \\  \bottomrule
\end{tabular}%\vspace{-2mm}
\end{table}

% For black-box tools ZOO~\cite{CZSYH17}, DBA~\cite{BRB18},
% NES-PGD~\cite{IEAL18} and Bandits~\cite{IEM2018PriorCB},
% the tuned parameters are given in Table~\ref{tab:Appendix2} for reproducing our results.
% These input parameters are related to the perturbation step size, constraints, confidence,
% and the number of iterations.

For greedy search based version of C\&W, the problem of searching adversarial examples
is reduced to the solving of the following optimization problem:
\[{\tt minimize} \ c\times f(x+\delta)+\|\delta\|_2\]
where $c$ is an input parameter, $f$ is a loss function, $x$ is the original image,
and $\delta$ is the perturbation.
We tune the input parameter $c$  (i.e., init const in their implementation)
and the number of iterations of the binary search which is used
to minimize distortion once an adversarial example is found.
These fine-tuned parameters could improve TSR well after combining with greedy search,.

\begin{table}[t]\small
\caption{Parameters and results of the greedy search based version of C\&W}
\label{tab:tunepara4}
\centering
\begin{tabular}{ccccc}
\multicolumn{5}{l}{C\&W-$\textbf{L}_2$~\cite{CW17b} attack on MNIST+original model}\\
\multicolumn{5}{l}{Version1 (default): init const=1e-3, binary search=9, no greedy search} \\
\multicolumn{5}{l}{Version2: init const=1e-3, binary search=9, greedy search} \\
\multicolumn{5}{l}{Version3: init const=1, binary search=3, no greedy search} \\
\multicolumn{5}{l}{Version4: init const=1, binary search=3, greedy search} \\
\toprule
Version     & SR & TSR & MSE & Search steps \\ %& ATC\\
\midrule
\textbf{1}  & 100\% &22.89\% &4.57e-03 & N/A    \\ %   & 0.24\\
\textbf{2}  & 100\% &99.67\% &4.6e-03  & $\infty$(200000) \\ %& 0.24+12.84\\
\textbf{3}  & 100\% &60\%    &4.54e-03 & N/A    \\ %   & 0.09\\
\textbf{4}  & 100\% &100\%   &4.56e-03 & 237   \\ %   & 0.09+0.01\\
\bottomrule
\end{tabular}

\end{table}